\PassOptionsToPackage{table}{xcolor} % xcolor 옵션 미리 전달

\documentclass[10pt,twocolumn,letterpaper]{article}

\usepackage[pagenumbers]{iccv} % To force page numbers, e.g. for an arXiv version
\usepackage{amsmath,graphicx}
\usepackage{adjustbox}
\usepackage{url}
\usepackage{array}
\usepackage{multirow}
\usepackage{amssymb}
\usepackage{pifont}

\usepackage{float} % here for H placement parameter
\usepackage{multirow}
\usepackage{booktabs}
\usepackage{makecell}
\usepackage{tikz}
\usepackage{tipa}
\usepackage{setspace}
\usepackage{caption}
\usepackage{lipsum}
\usepackage{xcolor}   % xcolor은 여기서 옵션이 이미 전달되어 있음
\usepackage{arydshln}
\usepackage{CJKutf8}

\definecolor{lightblue}{rgb}{0.93,0.95,1.0} % Define custom color

\definecolor{iccvblue}{rgb}{0.21,0.49,0.74}
\usepackage[pagebackref,breaklinks,colorlinks,allcolors=iccvblue]{hyperref}

\def\paperID{2162} % *** Enter the Paper ID here
\def\confName{ICCV}
\def\confYear{2025}

\title{Zero-AVSR: Zero-Shot Audio-Visual Speech Recognition with LLMs \\ by Learning Language-Agnostic Speech Representations}

\author{Jeong Hun Yeo$^{1,}\thanks{Equal Contribution. $^\dagger$Corresponding Author.}$ \quad Minsu Kim$^{1,*}$ \quad Chae Won Kim$^1$ \quad Stavros Petridis$^2$ \quad Yong Man Ro$^{1,\dagger}$ \\
$^1$KAIST \quad $^2$Imperial College London\\
{\tt\small \{sedne246, ms.k, ymro\}@kaist.ac.kr}
}

\begin{document}
\maketitle
\begin{abstract}
We explore a novel zero-shot Audio-Visual Speech Recognition (AVSR) framework, dubbed Zero-AVSR, which enables speech recognition in target languages without requiring any audio-visual speech data in those languages. Specifically, we introduce the Audio-Visual Speech Romanizer (AV-Romanizer), which learns language-agnostic speech representations by predicting Roman text. Then, by leveraging the strong multilingual modeling capabilities of Large Language Models (LLMs), we propose converting the predicted Roman text into language-specific graphemes, forming the proposed Cascaded Zero-AVSR. Taking it a step further, we explore a unified Zero-AVSR approach by directly integrating the audio-visual speech representations encoded by the AV-Romanizer into the LLM. This is achieved through finetuning the adapter and the LLM using our proposed multi-task learning scheme. To capture the wide spectrum of phonetic and linguistic diversity, we also introduce a Multilingual Audio-Visual Romanized Corpus (MARC) consisting of 2,916 hours of audio-visual speech data across 82 languages, along with transcriptions in both language-specific graphemes and Roman text. Extensive analysis and experiments confirm that the proposed Zero-AVSR framework has the potential to expand language support beyond the languages seen during the training of the AV-Romanizer. The code and models are available \href{https://github.com/JeongHun0716/zero-avsr}{online}.

\end{abstract}    
\vspace{-0.5cm}
\section{Introduction}
\vspace{-0.2cm}
Humans rely on multi-modal information for effective communication, combining verbal cues (\eg, spoken words), nonverbal signals (\eg, facial expressions, gestures), and paralinguistic auditory cues (\eg, tone of voice) to convey meaning effectively. Notably, the correlation between auditory speech and visual speech (\ie, lip movements) can significantly enhance speech comprehension, particularly in noisy environments. Leveraging this advantage, numerous studies~\cite{afouras2018deep, petridis2018end, makino2019recurrent, ma2021end, ren2021learning, shi2022learning, hong2022visual, serdyuk2022transformer, ma2023auto, hong2023watch, cappellazzo2024large, rouditchenko2024whisper, haliassos2024unified} have explored Audio-Visual Speech Recognition (AVSR) using deep learning techniques. AVSR can be viewed as a multi-modal fusion of two uni-modal systems; Auditory Speech Recognition (ASR)~\cite{amodei2016deep,kim2017joint,prabhavalkar2023end} and Visual Speech Recognition (VSR)~\cite{ma2022visual, haliassos2022jointly, kim2023lip, kim2024efficient, yeo2024visual, yeo2024visual2, haliassos2024braven}. Prior works have demonstrated that jointly modeling both audio and visual speech representations can significantly improve the speech recognition performance of each uni-modal system, particularly in challenging noisy environments.

Despite significant advancements, AVSR has primarily been developed and evaluated on English language corpora. To extend its capabilities to a wider range of languages, recent efforts have proposed multilingual audio-visual speech datasets~\cite{salesky2021multilingual, anwar2023muavic}, and have developed and evaluated multilingual AVSR methods~\cite{zinonos2023learning, han2024xlavs, burchi2024multilingual} on such databases. However, despite these performance improvements, existing approaches still face limitations in language expansion. This is because current multilingual AVSR datasets~\cite{salesky2021multilingual,anwar2023muavic} contain only nine languages, and it remains challenging to obtain a sufficient amount of labeled audio-visual data with transcriptions for diverse languages. To address this limitation, our main explorations include: 1) learning language-agnostic audio-visual speech representations, which facilitate easier language expansion, and 2) proposing a new multilingual audio-visual database that encompasses 82 languages, significantly expanding the linguistic diversity of existing datasets.

In this paper, we present a novel AVSR method that exhibits zero-shot language recognition ability\footnote{The `zero-shot' capability defined in this paper is limited to the speech modeling component. We assume that text data is available for all languages, whereas speech data is not accessible for some of them. Therefore, a `seen language' refers to a language that has been encountered during speech encoder training, whereas an `unseen language' refers to a language that is only available through text data, without any corresponding speech.}, \ie, a scenario in which no speech data in the target language is used at training. To achieve this, we aim to represent all languages in Roman text, converting language-specific graphemes into language-agnostic pronunciations. Here, we note that pre-trained Large Language Models (LLMs)~\cite{touvron2023llama,achiam2023gpt,jiang2023mistral} already possess knowledge of modeling this Roman-grapheme mapping, and propose to leverage this comprehensive multilingual processing ability of LLMs in our proposed zero-shot AVSR framework. Concretely, we propose an Audio-Visual Speech Romanizer (AV-Romanizer), which predicts Roman text (\ie, language-agnostic) from input multilingual audio-visual speech data. We then demonstrate that the proposed AV-Romanizer can be directly employed to achieve zero-shot AVSR by converting the predicted Roman text into language-specific graphemes using a pre-trained LLM, even when the target language is not used during the training of the AV-Romanizer. We refer to this usage case as Cascaded Zero-AVSR, where the proposed AV-Romanizer and a pre-trained LLM form a cascaded system.

Going one step further, we explore a model that holds considerable promise, namely Zero-AVSR, where the LLM is also fine-tuned to match the specific use case for zero-shot recognition, thereby improving performance. Specifically, Zero-AVSR is trained in a multi-task fashion composed of two tasks. 
The first task involves aligning the AV-Romanizer and LLM by fine-tuning an adapter and the LLM on \underline{seen languages}, enabling the audio-visual speech representation obtained from the AV-Romanizer to be seamlessly embedded into the learned text space of the LLM.
The second task focuses on learning to de-romanize, where the LLM is trained to convert Roman texts into language-specific graphemes using only text data for \underline{both seen and unseen languages}. Therefore, Zero-AVSR constructs knowledge connecting the audio-visual speech representation with the LLM's embedding space of seen languages at the first task, while extending the learned knowledge to more languages, including unseen languages, at the second task.

In order to train the proposed Zero-AVSR, a sufficient amount of audio-visual speech data is essential to capture the full spectrum of phonetic and linguistic diversity. In this context, we introduce a Multilingual Audio-Visual Romanized Corpus (MARC), which provides Roman transcriptions for approximately 2,916 hours of audio-visual data across 82 languages, along with their corresponding transcriptions in language-specific graphemes.

The contributions of this paper can be summarized as:
\begin{itemize}
    \item We explore Zero-AVSR, the first zero-shot AVSR framework, designed to operate in scenarios where no speech training data is available for the target language.
    \item We propose the MARC dataset, which comprises Roman transcriptions for 2,916 hours of audio-visual speech data across 82 languages.
    \item We explore two types of zero-shot AVSR frameworks, one of which is Cascaded Zero-AVSR, a framework that can be used with any type of LLM without fine-tuning, even in API form. The other is Zero-AVSR, an improved framework by finetuning the LLM with the proposed multi-task learning.
    %\item We comprehensively analyze Zero-AVSR's effectiveness across multiple factors, including the number of languages, language families, and dataset sizes.
\end{itemize}

\section{Related Work}
\subsection{Audio-Visual Speech Recognition}
Recently, AVSR has gained significant attention for its practical benefits, particularly in enhancing robust speech recognition in noisy environments. Along with the development of large-scale audio-visual datasets \cite{afouras2018lrs3, chung2017lip}, early work \cite{petridis2018end} introduced end-to-end AVSR frameworks based on Bidirectional Gated Recurrent Units (BGRUs). Subsequently, researchers improved these architectures by incorporating Transformer \cite{afouras2018deep, vaswani2017attention} and Conformer \cite{gulati2020conformer, ma2021end}, leading to notable performance gains. Concurrently, other researchers have explored multimodal learning strategies, including self-supervised learning \cite{shi2022learning, haliassos2022jointly, han2024xlavs}, leveraging knowledge from pre-trained ASR models in AVSR \cite{rouditchenko2024whisper, ma2023auto}, and harnessing the context-modeling capabilities of LLMs for speech recognition \cite{cappellazzo2024large, yeo2024visual}.

Despite these remarkable advances, most existing AVSR research has primarily focused on English. To address this gap, some studies have begun exploring the effectiveness of AVSR in multilingual contexts, leveraging newly introduced multilingual audio-visual databases~\cite{anwar2023muavic, salesky2021multilingual} spanning nine languages. Building on the progress made in English-based AVSR, recent multilingual AVSR approaches \cite{han2024xlavs, burchi2024multilingual} have expanded its progress into multilingual AVSR models. In particular, because obtaining labeled multilingual audio-visual data is challenging, self-supervised learning \cite{han2024xlavs} has shown notable promise for improving performance by leveraging abundant unlabeled multilingual audio-visual data. 

While these efforts have successfully extended AVSR’s effectiveness to nine languages, it remains challenging to expand its impact to additional languages. Unlike previous work, which mainly focused on improving performance on publicly available multilingual databases, we explore language expansion in multilingual AVSR by constructing a zero-shot AVSR framework. The proposed framework can recognize speech in the target language without requiring any speech training data for that specific language.

\begin{table*}[t!]
\renewcommand{\arraystretch}{1.2}
\renewcommand{\tabcolsep}{2.5mm}
\centering
\resizebox{0.8\linewidth}{!}{
\begin{tabular}{cccccccccccc}
\toprule
\multirow{2.5}{*}{\textbf{Romanizer}} 
& \multirow{2.5}{*}{\textbf{De-romanizer}} 
& \multicolumn{8}{c}{\textbf{Target Language (CER$\downarrow$)}} & \multirow{2.5}{*}{\makecell{\textbf{Avg} \\ \textbf{(w/o Eng)}}} \\
\cmidrule(lr){3-10}
 & & \textbf{Ara} & \textbf{Deu} & \textbf{Ell} & \textbf{Spa} & \textbf{Fra} & \textbf{Ita} & \textbf{Por} & \textbf{Rus} &   \\
\midrule
Llama3.1-70B~\cite{touvron2023llama} & Llama3.1-70B~\cite{touvron2023llama}  & 34.9 & 8.4 & 16.9 & 12.3 & 14.0 & 10.0 & 11.1 & 10.9 & 13.8 \\
Llama3.1-70B~\cite{touvron2023llama} & GPT-4o-mini~\cite{achiam2023gpt}  & 24.6 & 3.1 & 10.6 & 4.0 & 2.8 & 1.8 & 6.6 & 6.1 & 7.2 \\
Uroman~\cite{hermjakob2018out} & Llama3.1-70B~\cite{touvron2023llama}  & 15.0 & 4.1 & 5.8 & 9.8 & 4.2 & 4.9 & 4.4 & 14.8 & 7.4 \\
Uroman~\cite{hermjakob2018out} & GPT-4o-mini~\cite{achiam2023gpt} & 20.0 & 1.7 & 8.6 & 1.0 & 2.2& 1.3& 2.0 & 8.6 & 4.8 \\
GPT-4o-mini~\cite{achiam2023gpt} & Llama3.1-70B~\cite{touvron2023llama}  & 11.8 & 3.6 & 5.5 & 12.4 & 5.5 & 2.9 & 3.1 & 7.4 & 6.0  \\
\rowcolor{gray!10}GPT-4o-mini~\cite{achiam2023gpt} & GPT-4o-mini~\cite{achiam2023gpt}  & 13.1 & 1.9 & 7.1 & 0.7 & 2.5& 1.2& 1.9 & 3.9 & 4.0\\
% rowcolor, cellcolor
\bottomrule
\end{tabular}
}
\vspace{-0.2cm}
\caption{Reconstruction test results on MuAViC to evaluate the effectiveness of different methods in romanization and de-romanization.}
\label{tab:1}
\vspace{-0.5cm}
\end{table*}

\subsection{Zero-Shot Speech Recognition}
Recent research in speech recognition has advanced toward supporting multiple languages by employing effective methods such as self-supervised learning, which has been validated on English ASR. This rapid progress has been made possible by the use of large-scale multilingual audio datasets and carefully designed training methods~\cite{radford2023robust, pratap2024scaling}. However, obtaining a sufficient amount of labeled data with transcriptions for all languages remains a challenge.

To overcome this limitation, researchers have begun to explore zero-shot speech recognition. Early works~\cite{liu2018completely, chen2019completely, baevski2021unsupervised} proposed unsupervised ASR methods leveraging both unlabeled audio and text data. A more efficient approach that requires only unlabeled text was proposed by~\cite{li2022asr2k}. This method utilizes language-agnostic allophones, which are subsequently converted into language-specific phonemes~\cite{li2020universal}. Then, by using a grapheme-to-phoneme (G2P) system~\cite{li2022zero}, they generate words from phoneme sequences. More recently, ~\cite{zhao2024scaling} demonstrates that using a romanized form instead of phonemes offers a simpler yet effective alternative. 

Building on these findings, we also aim to develop a zero-shot AVSR framework by employing romanized text. Unlike \cite{zhao2024scaling}, which relies on language-specific language models to decode Roman text into language-specific graphemes, we demonstrate that a pre-trained Large Language Model (LLM) can effectively perform this task, thereby eliminating the need for training multiple language-specific language models.

\subsection{Speech Large Language Model}
LLMs have significantly impacted Natural Language Processing (NLP), as demonstrated by their ability to perform a wide range of tasks~\cite{radford2019language}. Building on this success, recent research efforts have begun exploring their effectiveness in other modalities~\cite{lakhotia2021generative, latif2023sparks, park2024lets}. In particular, integrating speech with LLMs has effectively leveraged these models' language understanding capabilities, leading to substantial improvements in the speech domain. To achieve this, simple and effective adaptation methods such as LoRA~\cite{hu2021lora} and a window-level Q-former~\cite{tang2023salmonn}, have been proposed to align audio-based speech with LLMs. Concurrently, these approaches have expanded the range of applications, enabling multi-task capabilities~\cite{chu2023qwen, tang2023salmonn} and multi-modal processing~\cite{cappellazzo2024large}. Similar to these research trends, we also explore a unified Zero-AVSR framework that fine-tunes a pre-trained LLM using QLoRA~\cite{dettmers2023qlora}, enabling it to directly accept and process encoded audio-visual speech features.

\section{Method}
We propose a novel multilingual AVSR framework, Zero-AVSR, that enables zero-shot audio-visual speech recognition even though speech training data for the target language is unavailable. Specifically, we propose the AV-Romanizer, which predicts language-agnostic speech representations (\ie, Roman) from input audio-visual speech. Subsequently, we propose to leverage LLMs to convert these representations back into language-specific graphemes.

\begin{figure*}[t]
\centering
\centerline{\includegraphics[width=15cm]{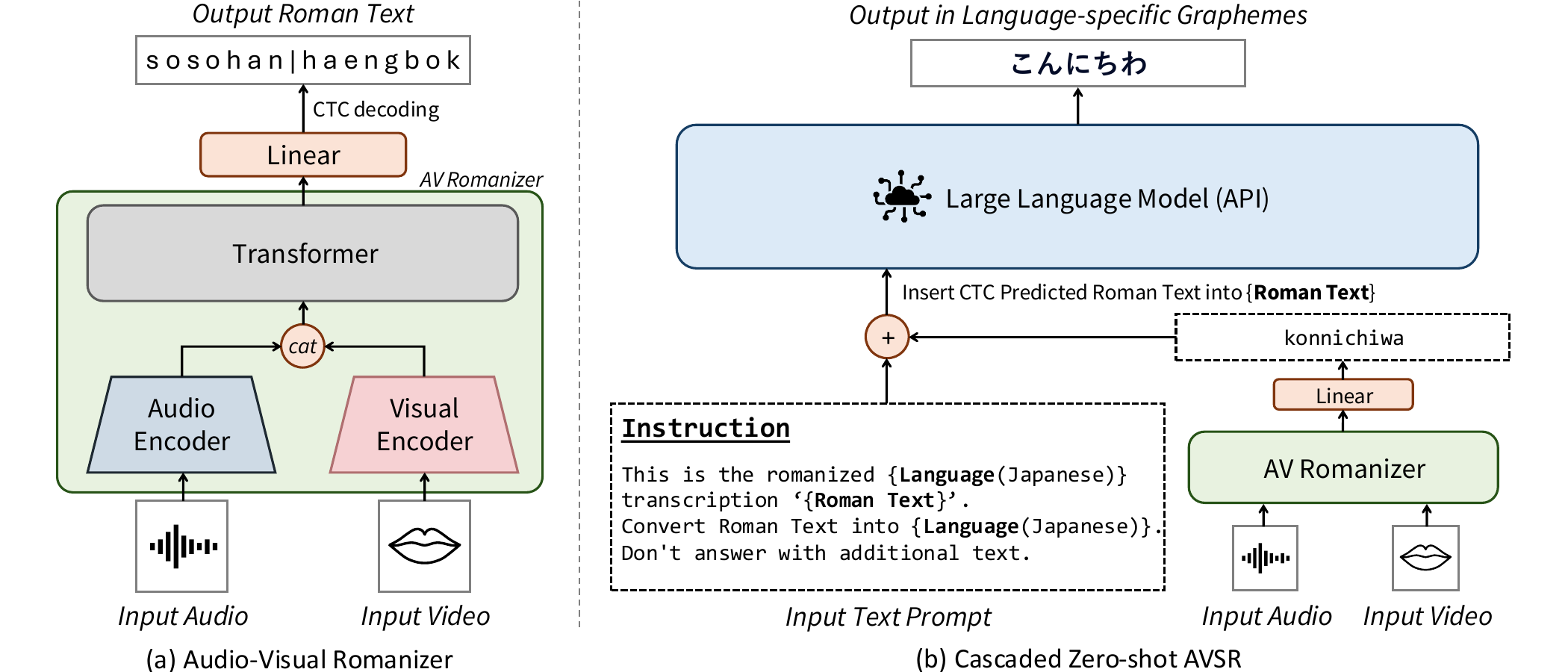}}
\vspace{-0.2cm}
\caption{Illustration of (a) Audio-Visual Speech Romanizer (AV-Romanizer): It is pre-trained to learn language-agnostic representations by predicting romanized text through CTC loss. (b) Cascaded Zero-shot AVSR: By providing instructions along with the predicted Roman text from the proposed AV-Romanizer, diverse LLMs can be employed to predict graphemes from the predicted Roman text. Please note that, as the AV-Romanizer has learned to generate pronunciation information (Roman text) from the input speech, it can convert even unseen languages into Roman text. Since LLMs already contain information about the target language (key assumption of this paper), they can then convert this Roman text into the target language.
}
\label{fig:1}
\vspace{-0.5cm}
\end{figure*}

\subsection{Multilingual Audio-Visual Romanized Corpus}
While it is known that different languages share common pronunciation features at the phoneme level~\cite{schultz2001language, vu2014multilingual,kim2024textless}, the current publicly available AVSR datasets~\cite{anwar2023muavic} may be insufficient for representing the phonemic diversity of diverse languages and for developing language-agnostic audio-visual speech representations. To address this limitation, we propose the Multilingual Audio-Visual Romanized Corpus (MARC). MARC is driven by mixing existing audio-visual corpora, LRS3~\cite{afouras2018lrs3} and MuAViC~\cite{anwar2023muavic}, and unlabeled audio-visual datasets, VoxCeleb2~\cite{chung2018voxceleb2} and AVSpeech~\cite{ephrat2018looking}. For the unlabeled datasets, we employ pre-trained language identification and ASR models~\cite{pratap2024scaling} to obtain the language ID and the language-specific graphemes for each data point, similar to~\cite{burchi2024multilingual, ma2023auto, yeo2024visual2}. During language identification, as the pre-trained model may misclassify the language, we filter out annotations with prediction probabilities below a predetermined threshold to reduce the errors. Finally, we convert the transcriptions of the above datasets into Roman text. 
 
Prior to this, we evaluate which method is most suitable for romanizing transcriptions between using a romanization tool~\cite{hermjakob2018out} and LLMs. To this end, we perform a reconstruction test, where the ground-truth transcriptions are transformed into Roman text by using either romanization tool or one of the LLMs, and then they are de-romanized into the original transcriptions. However, as romanization is not fully reversible, the de-romanization process is non-trivial. One may construct language-specific lexicons to de-romanize Roman text~\cite{zhao2024scaling}, but this approach cannot fully capture the complex few-to-many mapping of de-romanization. Instead, we employ LLMs as a de-romanizer and find that they already possess the capability to transform between Roman and graphemes freely. Table~\ref{tab:1} shows the reconstruction test results using different methods as a romanizer and de-romanizer on MuAViC dataset. Through the test, we found that GPT-4o-mini~\cite{achiam2023gpt} achieves the best performance when it is utilized as both romanizer and de-romanizer. Therefore, we romanize the transcriptions of the proposed MARC with GPT-4o-mini.

The resulting dataset, MARC, is composed of 82 languages, and includes 2,916 hours of audio-visual data, and text transcriptions in both language-specific graphemes and Roman text. Detailed information about the dataset can be found in Sec.~\ref{sec:6}.

\subsection{Audio-Visual Speech Romanizer}
If we train an acoustic model to predict the pronunciation of input speech instead of predicting language-specific graphemes, we can employ a fixed number of characters for diverse languages and potentially represent languages even not used during training. Previous approaches~\cite{li2022asr2k, li2020universal, li2022zero} have often relied on mapping allophones to phonemes. While these approaches can model language-agnostic representations, recent work~\cite{zhao2024scaling} demonstrates that using a romanized form offers a simpler yet effective alternative. We also adopt Roman text as our language-agnostic representation.

\begin{figure*}[t]
\centering
\centerline{\includegraphics[width=17cm]{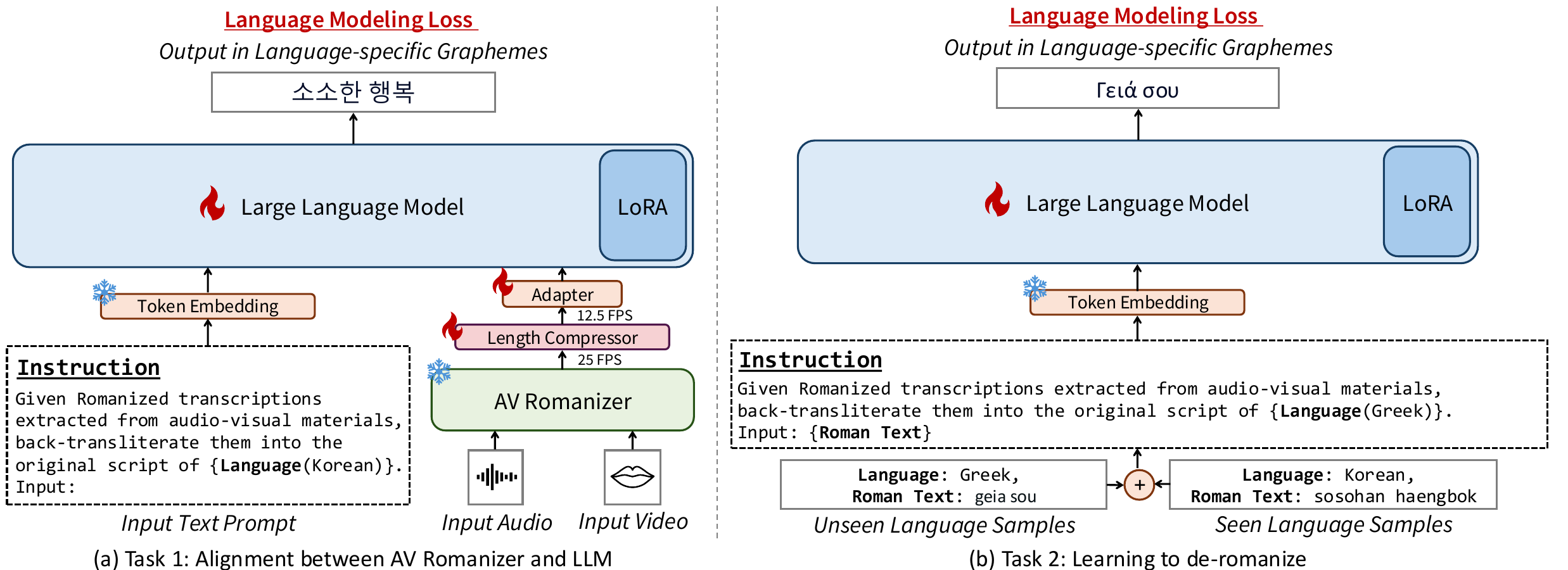}}
\vspace{-0.2cm}
\caption{Illustration of the proposed multi-tasks to train Zero-AVSR. (a) Task 1: Alignment between the AV Romanizer and a LLM. By using the paired audio-visual speech inputs and ground truth roman text, we align the audio-visual speech representations into LLM space (\ie, text). (b) Task 2: Learning to de-romanize. By leveraging abundant text-only data, we fine-tune the LLM to de-romanize input roman text into language-specific graphemes. This approach enables us to cover a wider range of languages, as text data is more readily available than audio-visual speech data for diverse languages.
}
\label{fig:2}
\vspace{-0.5cm}
\end{figure*}

The proposed Audio-Visual Speech Romanizer (AV-Romanizer) predicts Roman text for capturing pronunciation as it sounds from the input audio-visual speech, regardless of language, as shown in Fig.~\ref{fig:1}a. It comprises four main components: an audio encoder $\mathcal{F}_{a}$, a visual encoder $\mathcal{F}_{v}$, a transformer $\mathcal{B}$, and a linear layer for predicting Roman text. Given the training sample $(x_{a}, x_{v}, y)$, where $x_{a}$ represents the log-filterbank features extracted from the input audio, $x_{v}$ denotes the video capturing lip movements, and $y$ is the target Roman text, we first encode the audio-visual speech features. Specifically, the audio features $f_a = \mathcal{F}_{a}(x_a) \in \mathbb{R}^{T \times D}$ and the visual features $f_v= \mathcal{F}_{v}(x_v) \in \mathbb{R}^{T \times D}$ are extracted using the audio and visual encoders, respectively. Note that the lengths of audio and visual features are synchronized through pre-processing and strided convolution. Next, we concatenate the audio and visual features along the channel dimension and then these combined features are fed into a transformer encoder to encode the audio-visual speech features $f_{av}$. This process can be formulated as follows: $f_{av}= \mathcal{B}((f_{a} \oplus f_{v}) W)$, where $\oplus$ denotes concatenation along the channel dimension, and $W \in \mathbb{R}^{2D \times D}$ is a weight matrix of a linear layer applied to preserve the original dimensionality. Finally, a linear layer is applied to predict the output roman tokens $\hat{y}$. The proposed AV-Romanizer is trained with Connectionist Temporal Classification (CTC)~\cite{graves2006connectionist} objective function.

After training on sufficient data that covers a diverse range of phones, the AV-Romanizer can predict Roman text by capturing how input audio-visual speech is pronounced, even though the language was not used during training.

\subsection{Cascaded Zero-shot AVSR}
If we can transform the Roman text into language-specific graphemes, then the speech recognition process is complete. To construct the complete AVSR pipeline, we cascade the AV-Romanizer and an LLM. The AV-Romanizer transcribes input audio-visual speech into Roman text. Then, the predicted Roman text is de-romanized by a pre-trained LLM. Here, the zero-shot language recognition ability arises. Even though a language has not been used to train the AV-Romanizer, it can still predict how the input speech sounds like, in Roman text. By taking the predicted text, a pre-trained LLM can convert the text into the target language, which is within its knowledge. We named this cascaded form as Cascaded Zero-AVSR and depict in Fig.~\ref{fig:1}b.

Specifically, an instruction to convert the Roman text into the target language is incorporated along with the predicted Roman text from the AV-Romanizer. This text is then used as input for the LLMs as shown in Fig.~\ref{fig:1}b. Since this approach does not require finetuning the LLMs and the input is purely text-based, diverse LLMs can be employed for Cascaded Zero-AVSR, even in API form.

\subsection{Zero-shot AVSR with Multi-task Training}
Although we have seen the potential of our proposed AV-Romanizer to interact with LLMs at the text level as a cascaded system, recent works~\cite{tang2023salmonn, chu2023qwen} have demonstrated that systems that directly integrate speech representations with LLMs can achieve optimal performance. Motivated by this, we explore an unified model, namely Zero-AVSR, which directly integrates audio-visual speech representations with LLMs instead of using the predicted Roman text itself. To this end, Zero-AVSR is trained on two tasks to enable zero-shot language recognition scenarios. The first task aims to align the audio-visual speech features encoded by the AV-Romanizer with the text embeddings of LLMs. The second task involves learning to de-romanize languages, encompassing both seen and unseen languages.

\subsubsection{Task 1: Aligning the AV-Romanizer with LLMs}
In this task, our goal is to align the audio-visual speech representations obtained from the AV-Romanizer with the text embeddings of LLMs. Since audio-visual speech has a higher temporal resolution compared to text and thereby contains redundant information, a length compressor is typically employed to reduce the computational burden, especially when incorporating with LLMs~\cite{chen2024llmasr,yeo2024visual,hu2024large,cappellazzo2024large}.

As shown in Fig.~\ref{fig:2}a, given the audio-visual features $f_{av}$ extracted at the penultimate layer before the classification head of the AV-Romanizer, we first apply a length compressor to halve the length of the features. Then, an adapter maps the compressed audio-visual speech features into the LLM's embedding space. Finally, the embedded audio-visual speech features are concatenated with the text embedding of the instruction to form the input to the LLM. By setting the target as language-specific graphemes, the model is trained using a typical language modeling objective. Here, the pre-trained weights of the LLM including token embeddings and AV-Romanizer are kept frozen, while only the LoRA weights at the LLMs, length compressor, and the adapter are finetuned. As the task 1 requires audio-visual speech data, it can only be performed using seen languages. The knowledge for unseen languages will be extended in the task 2.

\begin{table*}[ht!]
\renewcommand{\arraystretch}{1.3}
\renewcommand{\tabcolsep}{1.5mm}
\centering
\resizebox{0.85\linewidth}{!}{
  \begin{tabular}{ccccccccccccccc}
    \toprule
    \multirow{2.5}{*}{\textbf{Method}} 
    & \multirow{2.5}{*}{\textbf{Modality}}
    & \multicolumn{2}{c}{\textbf{AV Training Hours}}
    & \multirow{2.5}{*}{\makecell{\textbf{Support} \\ \textbf{\# Langs}}} 
    & \multicolumn{9}{c}{\textbf{Target Language (WER(\%)$\downarrow$)}} 
    & \multirow{2.5}{*}{\makecell{\textbf{Avg} \\ \textbf{(w/ Eng)}}} 
    \\
    \cmidrule(lr){3-4} \cmidrule(lr){6-14}
    & 
    & \textbf{Unlabeled} & \textbf{Labeled}
    & 
    & \textbf{Ara} & \textbf{Deu} & \textbf{Ell} & \textbf{Spa} & \textbf{Fra} & \textbf{Ita} & \textbf{Por} & \textbf{Rus} & \textbf{Eng} &  \\
    \hline
    \rowcolor{gray!20}\multicolumn{15}{c}{\textbf{Non-Zero Shot Multilingual AVSR Models}} \\
    \multirow{1}{*}{AV-HuBERT \cite{shi2022learning}} 
        & AVSR & 1,759* & 1,200 & 9 & 89.4 & 52.0 & 46.2 & 17.4 & 20.3 & 20.8  & 22.1 & 44.7 & 1.7 & 35.0 \\
    \multirow{1}{*}{u-HuBERT \cite{hsu2022u}} 
        & AVSR & 1,759*(452*) & 1,200 & 9 & 89.3 & 52.1 & 46.4 & 17.3 & 20.5 & 21.2 & 21.9 & 44.4 & 1.9 & 35.0\\
    \multirow{1}{*}{XLAVS-R 300M \cite{han2024xlavs}} 
        & AVSR & 1,200(436K) & 1,200 & 9 & 81.7 & 44.7 & 24.3 & 10.9 & 14.4 & 12.8 & 13.2 & 32.7 & 2.4 & 26.3 \\
    \multirow{1}{*}{XLAVS-R 2B \cite{han2024xlavs}} 
        & AVSR & 1,200(436K) & 1,200 & 9 & 79.3 & 44.4 & 19.0 & 9.1 & 12.3 & 10.6 & 11.2 & 25.0 & 1.7 & 23.6\\ \hline
     \rowcolor{gray!20}\multicolumn{15}{c}{\textbf{Zero-Shot Multilingual ASR/AVSR Models}} \\
    MMS Zero-shot \cite{zhao2024scaling} & ASR & (436K) & (40K) & 1,078+ & 84.9 & 31.5 & 47.9 & 17.7 & 33.6 & 19.0 & 35.5 & 42.8 & 35.7 & 38.9 \\ \hdashline
    \multirow{1}{*}{\makecell{\textbf{Cascaded Zero-AVSR}}} 
        & AVSR & 1,759* & 2,916 & 82+ & 82.1 & 29.3 & 47.2 & 16.3  & 28.9 & 21.6 & 20.2 & 42.9 & 2.9 & 30.2 \\
    \multirow{1}{*}{\textbf{Zero-AVSR}}
         & AVSR & 1,759* & 2,916 & 82+ & 81.4 & 27.8 & 38.4 & 13.1 & 14.3 & 15.9 & 15.4 & 32.6 & 1.5 & 25.2 \\
    \hline

\end{tabular}}
\vspace{-0.2cm}
\caption{Comparisons with state-of-the-art methods on MuAViC. Note that an asterisk (*) denotes the use of English-only data, and values in parentheses indicate the amount of audio-only data employed.}
\vspace{-0.5cm}
\label{tab:2}
\end{table*}

\subsubsection{Task 2: Learning to De-romanize}
In this task, our goal is to train LLMs to de-romanize diverse languages. As the task 1 is performed using only seen languages, task 2 is essential to prevent LLMs from forgetting their multilingual ability which is the key for zero-shot language recognition. Specifically, as shown in Fig.~\ref{fig:2}b, the input is set to purely text, consisting of the instruction and the Roman text of all languages, including both seen and unseen languages. The target output is then set to language-specific graphemes, constructing a transformation task from Roman text to language-specific graphemes. During task 2 training, only the LoRA weights of the LLM are finetuned.

\section{Experimental Setup}
\subsection{Dataset}
\noindent{\bf Multilingual Audio-visual Romanized Corpus (MARC)} is the proposed dataset designed for zero-shot audio-visual speech recognition by labeling and integrating data from four existing datasets: LRS3~\cite{afouras2018lrs3}, MuAViC~\cite{anwar2023muavic}, VoxCeleb2~\cite{chung2018voxceleb2}, and AVSpeech~\cite{ephrat2018looking}. Details for each source dataset and MARC can be found in Sec.~\ref{sec:6}.

\subsection{Implementation details}
\textbf{Pre-processing.} We resample all video and audio to 25 fps and 16 kHz, respectively. We extract facial landmarks using the ReinaFace detector~\cite{ma2023auto} and crop the mouth region to a size of $96 \times 96$. For text data, we apply fairseq's text normalization. Finally, for data augmentation, we perform random cropping into a size of $88 \times 88$ and horizontal flipping following~\cite{ma2021end,shi2022learning} during all training processes.

\noindent \textbf{Architecture.} For the AV-Romanizer, we adopt the AV-HuBERT~\cite{shi2022learning} architecture, which comprises a visual encoder, an audio encoder, an audio-visual fusion module, and a transformer encoder. Additionally, we employ a linear projection layer for predicting Roman text. The visual encoder uses a ResNet-18 with a 3D convolution layer, while the audio encoder consists of a single linear layer. The transformer encoder, composed of 24 layers, features a model dimension of 1024, a feed-forward dimension of 4096, and 16 attention heads. For the length compressor, a 1D convolution with a kernel size of 2 and a stride of 2 is utilized. For Zero-AVSR, we employ Llama3.2-3B~\cite{touvron2023llama} as the decoder and finetune it using QLoRA~\cite{hu2021lora,dettmers2023qlora}, while GPT-4o-mini~\cite{achiam2023gpt} is employed for Cascaded Zero-AVSR.

\noindent \textbf{Training and Evaluation.} For training the AV-Romanizer, we employ a three-stage scheduler with 10K warmup steps, 40K hold steps, and 50K decay steps, along with a peak learning rate of 1e-4. Training is performed on 8 RTX 3090 GPUs with gradient accumulation set to 8. Also, the audio is randomly perturbed with acoustic noise sampled from MUSAN~\cite{snyder2015musan} with 0\,dB SNR. For training the Zero-AVSR with an LLM, a cosine scheduler is employed with 0.5K warmup steps and 29.5K decay steps, using 7 A6000 GPUs with gradient accumulation set to 9. After training Zero-AVSR, we evaluate its performance using beam search with a width of 2 and a temperature of 0.3.

\section{Experimental Results}
\vspace{-0.1cm}
\subsection{Comparison with the state-of-the-art methods}
\vspace{-0.1cm}
Although our primary goal is language expansion, it is also crucial to ensure that the proposed methods perform on seen languages at a level comparable to state-of-the-art approaches for real-world applications. To validate this, we first evaluate the effectiveness of our proposed methods, Cascaded Zero-AVSR and Zero-AVSR, which are trained using audio-visual data from all languages in MARC. Table~\ref{tab:2} summarizes the performance of our methods compared to several state-of-the-art approaches on the MuAViC~\cite{anwar2023muavic} dataset. We organize the comparison into two groups: (i) multilingual audio-visual speech recognition models that support a fixed set of nine languages, and (ii) methods that operate using Roman characters and cover a wider range of languages.

Among the multilingual models (Non-zero shot), both AV-HuBERT~\cite{shi2022learning} and u-HuBERT~\cite{hsu2022u} achieve an average WER of 35.0\% across nine target languages. In contrast, XLAVS-R~\cite{han2024xlavs} models highlight the benefits of increased model capacity and the use of unlabeled multilingual audio-visual data, along with large-scale audio-only data (436K hours) during pre-training. Their 300M and 2B variants achieve average WERs of 26.3\% and 23.6\%, respectively. We observe high WERs on Arabic (Ara) for all methods.

The audio-only zero-shot model, MMS Zero-shot~\cite{zhao2024scaling}, leverages 436K hours of unlabeled audio data for pre-training and then fine-tunes on 40K hours of labeled data spanning 1,078 languages. As a result, it supports over 1,078 languages but exhibits a relatively high average WER of 38.9\%. By learning language-agnostic audio-visual speech representations and integrating the multilingual language understanding capabilities of LLMs, the proposed methods, Cascaded Zero-AVSR and Zero-AVSR, outperform the previous approach, achieving average WERs of 30.2\% and 25.2\%, respectively. Compared to existing multilingual AVSR approaches that support only 9 languages, the proposed methods significantly expand AVSR capabilities to a broader range of languages by training on the proposed MARC, while achieving the comparable performances. In addition, Zero-AVSR achieves state-of-the-art performances in English (Eng) and German (Deu), with WERs of 1.5\% and 27.8\%, respectively.

\begin{table}[t!]
\renewcommand{\arraystretch}{1.3}
\renewcommand{\tabcolsep}{1mm}
\centering
\resizebox{0.999\linewidth}{!}{
\begin{tabular}{ccccccccccc}
\toprule
\multirow{2.5}{*}{\textbf{Method}} 
& \multirow{2.5}{*}{\makecell{\textbf{Unseen} \\ \textbf{Lang.}}} 
& \multicolumn{8}{c}{\textbf{Target Language (CER(\%)$\downarrow$)}}
& \multirow{2.5}{*}{\makecell{\textbf{Avg} \\ \textbf{(w/o Eng)}}}\\
\cmidrule(lr){3-10}
& & \textbf{Ara} & \textbf{Deu} & \textbf{Ell} & \textbf{Spa} & \textbf{Fra} & \textbf{Ita} & \textbf{Por} & \textbf{Rus} & \\
\midrule
\multirow{9}{*}{\makecell{\textbf{Cascaded} \\ \textbf{Zero-AVSR} \\ (GPT-4o-mini)}}
  & Ara   & \cellcolor{blue!30}67.6 & 17.1  & 23.2 & 8.3 & 14.3 & 10.2 & 10.0 & 20.6 & 20.8 \\
  & Deu   & 53.7 & \cellcolor{blue!30} 46.9 & 22.3 & 8.2 & 13.7 & 9.7  & 9.8  & 20.2 & 25.8 \\
  & Ell   & 55.6 & 17.0 & \cellcolor{blue!30} 45.3 & 8.1 & 14.5 & 10.0 & 9.6  & 21.2 & 23.5 \\
  & Spa   & 54.8 & 17.2 & 22.5 & \cellcolor{blue!30}19.0 & 14.3 & 10.3 & 10.1 & 20.5 & 20.4 \\
  & Fra   & 55.2 & 17.2 & 22.0 & 8.4 & \cellcolor{blue!30} 44.6 & 9.9  & 10.0 & 20.6 & 21.4 \\
  & Ita   & 54.2 & 17.5 & 22.5 & 8.3 & 13.7 & \cellcolor{blue!30} 23.9 &  10.0 & 20.9 & 20.9 \\
  & Por   & 54.4 & 17.1 & 22.6 & 8.3 & 14.2 & 10.1 & \cellcolor{blue!30}37.7 & 21.0 & 22.9 \\
  & Rus   & 55.2 & 17.1 & 22.6 & 8.4 & 14.0 & 10.1 & 9.9 & \cellcolor{blue!30}40.0 & 21.9 \\
\hline
\multirow{9}{*}{\makecell{\textbf{Zero-AVSR} \\ (Llama3.2-3B)}}
  & Ara   & \cellcolor{blue!30}76.5 & 16.2 & 21.6 & 6.8 & 7.4 & 7.0 & 7.7 & 18.5 & 19.7 \\
  & Deu   & 56.9 & \cellcolor{blue!30} 52.9 & 22.4 & 7.4 & 8.2 & 6.9 & 7.7 & 18.7 & 26.4 \\
  & Ell   & 53.9 & 16.1 & \cellcolor{blue!30} 62.1 & 6.8 & 8.1 & 6.8 & 7.8 & 18.4 & 24.3 \\
  & Spa   & 57.8 & 16.7 & 24.4 & \cellcolor{blue!30}19.7 & 8.5 & 7.5 & 8.6 & 20.2 & 19.0 \\
  & Fra   & 55.3 & 16.0 & 21.2 & 6.9 & \cellcolor{blue!30}54.6 & 7.2 & 7.9 & 18.1 & 20.7 \\
  & Ita   & 61.6 & 17.2 & 22.6 & 7.1 & 8.1 &  \cellcolor{blue!30} 25.1 & 7.7 & 18.6 & 20.7 \\
  & Por   & 58.0 & 16.3 & 24.7 & 7.5 & 7.7 & 7.4 & \cellcolor{blue!30}44.0 & 18.8 & 22.3 \\
  & Rus   & 57.7 & 16.0 & 22.7 & 7.1 & 7.7 & 6.9 & 7.6 & \cellcolor{blue!30}45.4 & 21.5 \\

    \bottomrule
  \end{tabular}}
\vspace{-0.2cm}
\caption{The zero-shot language AVSR performances of Cascaded Zero-AVSR and Zero-AVSR on MuAViC. We train 8 AV-Romanizers, setting each language as an unseen language, and evaluate their zero-shot performance, shown in blue-colored cells.}
\vspace{-0.5cm}
\label{tab:3}
\end{table}

\subsection{Effectiveness on Unseen Languages}
To validate the effectiveness of the proposed methods on unseen languages, we train eight different models for each of Cascaded Zero-AVSR and Zero-AVSR. In each experiment, we train the model by excluding the audio-visual data of the target unseen language. For example, if Spanish (Spa) is the target unseen language, the AV-Romanizer is trained without using any Spanish data and then evaluated on 8 languages, including both seen and unseen languages. Given the dominant status of English, we exclude its results from our analysis to focus on other relatively low-resource languages, allowing us to better understand the effectiveness of Zero-AVSR in zero-shot speech recognition.

For the Cascaded Zero-AVSR, we first predict the Roman text by using the AV-Romanizer and then de-romanize it using GPT-4o-mini~\cite{achiam2023gpt}. The zero-shot AVSR performances (CER) of Cascaded Zero-AVSR for Arabic (Ara), German (Deu), Greek (Ell), Spanish (Spa), French (Fra), Italian (Ita), Portuguese (Por), and Russian (Rus) are 67.6\%, 46.9\%, 45.3\%, 19.0\%, 44.6\%, 23.9\%, 37.7\%, and 40.0\%, respectively. This indicates that even without training on any speech data, the Cascaded Zero-AVSR model can still predict transcriptions in unseen languages. For instance, the Cascaded Zero-AVSR can achieve a 62.3\% character-level accuracy when predicting Portuguese (Por) speech.

For the Zero-AVSR, we employ Llama3.2-3B~\cite{touvron2023llama}, a state-of-the-art open-source LLM, as the decoder. The zero-shot AVSR performances (CER) for the eight languages are as follows: Ara 76.5\%, Deu 52.9\%, Ell 62.1\%, Spa 19.7\%, Fra 54.6\%, Ita 25.1\%, Por 44.0\%, and Rus 45.4\%. Although the Cascaded Zero-AVSR and Zero-AVSR models are not directly comparable due to their different LLMs, we can confirm that incorporating speech features into the LLM and finetuning the model (\ie, Zero-AVSR) significantly improves performance on seen languages. Furthermore, zero-shot AVSR performance is also substantially improved when compared to the Cascaded Zero-AVSR using the same LLM, Llama3.2-3B, whose results can be found in Sec.~\ref{sec:5.3.4}. This result highlights the potential of the Zero-AVSR; If a better LLM is employed, its performance can be significantly enhanced.

\begin{table}[t!]
  \label{tab:word_level}
  \centering
  \resizebox{0.999\linewidth}{!}{
  \begin{tabular}{cccccc}
    \toprule
    \multirow{3}{*}{\textbf{Train Dataset}}
    & \multirow{3}{*}{\makecell{\textbf{\# Train} \\ \textbf{Language}}} 
    & \multirow{3}{*}{\makecell{\textbf{\# Train} \\ \textbf{Hours}}} 
    & \multirow{3}{*}{\makecell{\textbf{Unseen} \\ \textbf{Language}}} 
    & \multicolumn{2}{c}{\textbf{CER(\%)$\downarrow$}} \\
    \cmidrule(lr){5-6}
    & & & & \textbf{Zero-shot} & \makecell{\textbf{Avg} \\ \textbf{(w/o Eng)}}  \\
    \midrule
MuAViC \cite{anwar2023muavic} & 8 & 745 & Rus & 62.3 & 48.3 \\
\hdashline
\multirow{3}{*}{+ MARC} 
    & 8 & 1,944 & Rus & 61.0 & 28.3 \\
    & 40 & 2,418 & Rus & 49.5 & 25.1 \\
    & 81 & 2,793 & Rus & 40.0 & 21.9 \\
\midrule
MuAViC \cite{anwar2023muavic} & 8 & 724 & Ita & 34.8 & 49.0 \\
\hdashline
\multirow{3}{*}{+ MARC} 
    & 8 & 1,921 & Ita & 28.5 & 25.9 \\
    & 40 & 2,395 & Ita & 26.0 & 23.7 \\
    & 81 & 2,770 & Ita & 23.9 & 20.9 \\
\midrule
MuAViC \cite{anwar2023muavic} & 8 & 698 & Spa & 35.5 & 46.3 \\
\hdashline
\multirow{3}{*}{+ MARC} 
    & 8 & 1,851 & Spa & 29.9 & 25.4 \\
    & 40 & 2,326 & Spa & 20.5 & 22.6 \\
    & 81 & 2,700 & Spa & 19.0 & 20.4 \\
    
    \bottomrule
  \end{tabular}}
  \vspace{-0.2cm}
  \caption{Impact of the proposed MARC on Cascaded Zero-AVSR, evaluated by varying data amount and number of languages.}
  \label{tab:4}
  \vspace{-0.2cm}
\end{table}

\subsection{Ablation study}
\subsubsection{Effectiveness of the MARC Dataset}
To validate the effectiveness of the proposed MARC dataset, we conducted a gradual data addition experiment starting with the MuAViC portion of the MARC dataset. Specifically, we conduct four experiments for each of three languages, Russian (Rus), Italian (Ita), and Spanish (Spa), totaling 12 experiments. We measure the performance of the Cascaded Zero-AVSR and each language is treated as a target unseen language. The results are shown in Table~\ref{tab:4}.

First, we train the AV-Romanizer using only MuAViC portion of the MARC dataset, and achieve average CERs (\ie, including both unseen and seen languages) of 48.3\%, 49.0\%, and 46.3\% for Rus, Ita, and Spa, respectively. Next, by incorporating the remaining audio-visual data of the MARC with the same 8 languages in the MuAViC portion, we achieve significantly improved average CERs of 28.3\%, 25.9\%, and 25.4\% for Rus, Ita, and Spa, respectively. The results demonstrate the effectiveness of using MARC dataset and show that incorporating more data is beneficial for improving AVSR performance.

Then, to assess the impact of language diversity on modeling language-agnostic audio-visual speech representations, we train the AV-Romanizer using audio-visual speech data from 40 and 81 languages in the MARC dataset. That means the model is trained on more than just the 8 evaluation languages. The results show that incorporating data from 41 languages improves the performance for both unseen (\ie, zero-shot) and seen languages. Especially, the zero-shot performance for Russian is greatly improved from 61.0\% CER to 49.5\% CER. This suggests that incorporating a wider range of diverse phonetic information through the use of multiple languages can significantly enhance zero-shot speech recognition performance. Similarly, employing a total of 81 languages further improves both the zero-shot performance and the average performance. Again, this result confirms the importance of employing diverse languages and sufficient phonetic coverage in learning language-agnostic speech representations.

\subsubsection{Ablation Study using Different LLM Models}
As the Cascaded Zero-AVSR does not finetune the LLM, its performance is heavily dependent on the choice of LLM used, as expected. To investigate the performance of Cascaded Zero-AVSR according to different types of LLMs, we perform ablation study by employing 5 different LLMs. The AV-Romanizer is trained on all 82 languages. Table~\ref{tab:5} shows the ablation results. We can confirm that employing a larger model, which has a better ability to transform between Roman and language-specific graphemes, yields better performance. Notably, GPT-4o-mini~\cite{achiam2023gpt} achieves the best performance among the baselines. Therefore, we employ GPT-4o-mini as the de-romanizer for other Cascaded Zero-AVSR experiments. Please note that we employ Llama3.2-3B~\cite{touvron2023llama} for training Zero-AVSR as it has a feasible number of parameters.

\begin{table}[t!]
\renewcommand{\arraystretch}{1.3}
\renewcommand{\tabcolsep}{1.5mm}
\centering
\resizebox{0.999\linewidth}{!}{
% all model - CER
  \begin{tabular}{lccccccccc}
    \toprule
    \multirow{2.5}{*}{\textbf{Method}} 
    & \multicolumn{8}{c}{\textbf{Target Language (CER(\%)$\downarrow$)}} & \multirow{2.5}{*}{\makecell{\textbf{Avg} \\ \textbf{(w/o Eng)}}} \\
    \cmidrule(lr){2-9}
    & \textbf{Ara} & \textbf{Deu} & \textbf{Ell} & \textbf{Spa} & \textbf{Fra} & \textbf{Ita} & \textbf{Por} & \textbf{Rus} \\
    \midrule

Llama3.2-3B~\cite{touvron2023llama}  & 69.3 & 24.2 & 56.6 & 13.9 & 27.9 & 17.5 & 14.9 & 56.5 & 35.7 \\
Mistral-7B~\cite{jiang2023mistral}   & 71.2 & 26.7 & 32.8 & 13.5 & 23.9 & 16.2 & 17.7 & 36.5 & 29.6 \\
Llama3.1-8B~\cite{touvron2023llama}  & 61.4  & 21.7 & 39.4 & 14.8 & 20.4 & 14.1 & 14.4 & 30.9 & 27.3 \\
Llama3.1-70B~\cite{touvron2023llama}  & 56.1 & 18.1 & 26.7 & 9.5 & 15.6 & 11.1 & 11.1 & 23.3 & 21.3 \\
\rowcolor{gray!10}GPT-4o-mini~\cite{achiam2023gpt}  & 54.0 & 17.3 & 22.5 & 8.1  & 14.2 & 10.4  & 9.9 & 21.2 & 19.5 \\
    \bottomrule
  \end{tabular}}
\vspace{-0.2cm}
\caption{Performances of Cascaded Zero-AVSR using different types of LLMs. The AV-Romanizer is trained on all 82 languages.}
\label{tab:5}
\vspace{-0.5cm}
\end{table}

\subsubsection{Ablation Study on the Impact of Language Family}
Languages within the same language family~\cite{katzner2002languages} often exhibit similar grammatical structures and pronunciation patterns. Consequently, it is reasonable to expect that incorporating more data from languages within the same group as the target unseen language will enhance zero-shot speech recognition performance. To evaluate this hypothesis, we conduct ablation studies starting with a baseline model trained without data from the Romance family, the Turkic family, and Korean. We then incrementally add data to the baseline model and build two variants: one incorporating Spanish and Italian (Romance languages) and another adding Turkish and Korean (non-Romance languages), ensuring that the same amount of data is added to each. By comparing these two variants, we can assess the impact of linguistic similarity on zero-shot speech recognition performance. We report and analyze the zero-shot AVSR performances on Romance languages, including Spanish (Spa), Italian (Ita), French (Fra), and Portuguese (Por).

The ablation results are shown in Table~\ref{tab:6}. The first row shows the zero-shot AVSR performance of the baseline model. By gradually adding non-Romance languages, Turkish and Korean sequentially, the zero-shot performances are slightly improved for Spa and Ita, while there is no improvement for Fra and Por. In contrast, we can confirm that the zero-shot speech recognition performances for Romance languages are greatly improved by adding Spa and Ita, compared to adding the non-Romance languages. When Spa is added, the Ita and Por performances are improved to 28.1\% and 44.0\% CERs, respectively, while the performance of Fra remains unchanged. Furthermore, adding Italian data on top of the Spa added model results in an additional improvement, ultimately achieving 51.3\% and 41.9\% CERs for Fra and Por. These findings provide evidence that incorporating languages similar to the target language is more effective in zero-shot speech recognition.

\begin{table}[t!]
  \renewcommand{\arraystretch}{1.3}
  \renewcommand{\tabcolsep}{1.5mm}
  \centering
  \resizebox{0.999\linewidth}{!}{
  \begin{tabular}{lcccccc}
    \toprule
    \multirow{2.5}{*}{\textbf{Train Dataset}}
    & \multirow{2.5}{*}{\makecell{\textbf{\# Train} \\ \textbf{Language}}} 
    & \multirow{2.5}{*}{\makecell{\textbf{Unseen} \\ \textbf{Languages}}} 
    & \multicolumn{4}{c}{\textbf{CER(\%)$\downarrow$}} \\
    \cmidrule(lr){4-7}
    & & &  \textbf{Spa} & \textbf{Ita} & \textbf{Fra} & \textbf{Por}  \\
    \hline
Baseline & 67 & Spa, Ita, Fra, Por & 40.1 & 34.2 & 55.8 & 48.0  \\
\hdashline
\multicolumn{5}{l}{$\bullet$ \textbf{Adding Data from Different Language Family}} \\
\quad + Turkish (51 hrs)  & 68 & Spa, Ita, Fra, Por  & 38.1 & 33.6 & 61.0 & 48.2  \\

\quad \quad + Korean (129 hrs) & 69 & Spa, Ita, Fra, Por  & 38.4 & 33.6 & 59.7 & 49.5 \\
\hdashline
\multicolumn{5}{l}{$\bullet$ \textbf{Adding Data from Romance Subgroup}} \\
\quad + Spanish (51 hrs)  & 68 &  Ita, Fra, Por  & 11.5 & 28.1 & 56.6 & 44.0  \\

\quad \quad + Italian (129 hrs) & 69 & Fra, Por & 11.0 & 11.1 & 51.3 & 41.9 \\

\bottomrule 
  \end{tabular}}
\vspace{-0.2cm}
\caption{Ablation study on the impact of using data from the same language family (Romance) on zero-shot speech recognition.}
\vspace{-0.5cm}
\label{tab:6}
\end{table}

\subsubsection{The Effectiveness of Zero-AVSR}
\label{sec:5.3.4}
In order to confirm the effectiveness of the Zero-AVSR compared to the Cascaded Zero-AVSR, we also report the zero-shot language AVSR performance of Cascaded Zero-AVSR using Llama3.2-3B, in Table~\ref{tab:7}. Therefore, since Cascaded Zero-AVSR and Zero-AVSR utilize the same LLM here, we can evaluate the effectiveness of finetuning the LLM by employing the proposed multi-task framework. By comparing the zero-shot speech recognition performances (\ie, shown in blue-colored cells), we can confirm that Zero-AVSR improves performance over all 8 languages, demonstrating the effectiveness of incorporating speech features directly into the LLM instead of employing a text formula. Furthermore, when comparing the average CER, taking into account both seen and unseen languages, Zero-AVSR outperforms the Cascaded Zero-AVSR model. These results show the promise of Zero-AVSR that if a better LLM is employed and finetuned, performance can be improved even more.

\begin{table}[t!]
\renewcommand{\arraystretch}{1.3}
\renewcommand{\tabcolsep}{1mm}
\centering
\resizebox{0.999\linewidth}{!}{
\begin{tabular}{ccccccccccc}
\toprule
\multirow{2.5}{*}{\textbf{Method}} 
& \multirow{2.5}{*}{\makecell{\textbf{Unseen} \\ \textbf{Lang.}}} 
& \multicolumn{8}{c}{\textbf{Target Language (CER(\%)$\downarrow$)}}
& \multirow{2.5}{*}{\makecell{\textbf{Avg} \\ \textbf{(w/o Eng)}}}\\
\cmidrule(lr){3-10}
& & \textbf{Ara} & \textbf{Deu} & \textbf{Ell} & \textbf{Spa} & \textbf{Fra} & \textbf{Ita} & \textbf{Por} & \textbf{Rus} & \\
\midrule
\multirow{9}{*}{\makecell{\textbf{Cascaded} \\ \textbf{Zero-AVSR} \\ (Llama3.2-3B)}}
  & Ara   & \cellcolor{blue!30}86.6 & 25.2 & 56.3 & 13.5 & 33.4 & 15.5 & 18.5 & 55.3 & 37.9 \\
  & Deu   & 76.8 & \cellcolor{blue!30} 67.0 & 55.0 & 12.4 & 21.5 & 13.6  & 14.8  & 53.1 & 44.3 \\
  & Ell   & 68.5 & 25.5 & \cellcolor{blue!30} 75.2 & 13.9 & 29.2 & 15.3 & 14.8  &  58.8 & 39.3 \\
  & Spa   & 73.6 & 25.0 & 53.8  & \cellcolor{blue!30} 34.8 & 20.3 & 18.8 & 16.1 & 53.9 & 37.1  \\
  & Fra   & 76.8  & 24.7 & 56.7 & 15.5 & \cellcolor{blue!30} 82.2 & 14.7  & 17.6 & 54.3 & 40.0 \\
  & Ita   & 73.7 & 25.2 & 54.2 & 12.7 & 26.4 & \cellcolor{blue!30} 36.7 & 14.8  & 55.9 & 37.6 \\
  & Por   & 74.5 & 25.7 & 54.1 & 17.1 & 21.4 & 14.4 & \cellcolor{blue!30} 63.0 & 55.2  & 41.4 \\
  & Rus   & 68.6 & 24.6 & 55.8 & 14.1 & 22.2 & 16.8 & 14.6 & \cellcolor{blue!30} 79.8 & 38.1 \\
\hline
\multirow{9}{*}{\makecell{\textbf{Zero-AVSR} \\ (Llama3.2-3B)}}
  & Ara   & \cellcolor{blue!30}76.5 & 16.2 & 21.6 & 6.8 & 7.4 & 7.0 & 7.7 & 18.5 & 19.7 \\
  & Deu   & 56.9 & \cellcolor{blue!30} 52.9 & 22.4 & 7.4 & 8.2 & 6.9 & 7.7 & 18.7 & 26.4 \\
  & Ell   & 53.9 & 16.1 & \cellcolor{blue!30} 62.1 & 6.8 & 8.1 & 6.8 & 7.8 & 18.4 & 24.3 \\
  & Spa   & 57.8 & 16.7 & 24.4 & \cellcolor{blue!30}19.7 & 8.5 & 7.5 & 8.6 & 20.2 & 19.0 \\
  & Fra   & 55.3 & 16.0 & 21.2 & 6.9 & \cellcolor{blue!30}54.6 & 7.2 & 7.9 & 18.1 & 20.7 \\
  & Ita   & 61.6 & 17.2 & 22.6 & 7.1 & 8.1 &  \cellcolor{blue!30} 25.1 & 7.7 & 18.6 & 20.7 \\
  & Por   & 58.0 & 16.3 & 24.7 & 7.5 & 7.7 & 7.4 & \cellcolor{blue!30}44.0 & 18.8 & 22.3 \\
  & Rus   & 57.7 & 16.0 & 22.7 & 7.1 & 7.7 & 6.9 & 7.6 & \cellcolor{blue!30}45.4 & 21.5 \\

    \bottomrule
  \end{tabular}}
\vspace{-0.2cm}
\caption{The zero-shot language AVSR performances of Cascaded Zero-AVSR and Zero-AVSR using the same LLM (Llama3.2-3B) on MuAViC dataset. We train 8 AV-Romanizers, setting each language as an unseen language, and evaluate their zero-shot performance, which are shown in blue-colored cells.}
\vspace{-0.3cm}
\label{tab:7}
\end{table}

\begin{figure}[t]
\centering
\centerline{\includegraphics[width=8.5cm]{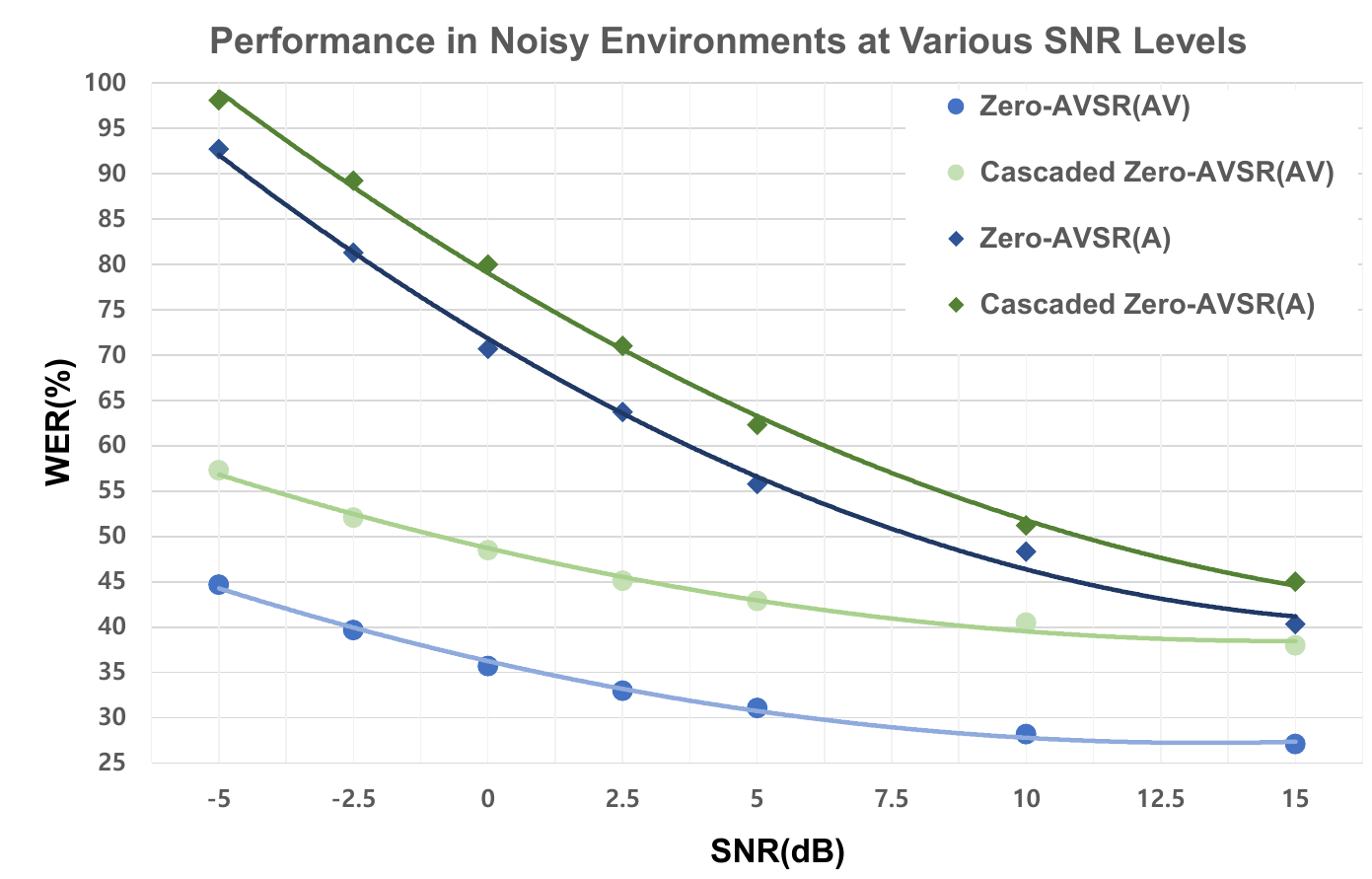}}
\vspace{-0.2cm}
\caption{The performances of Cascaded Zero-AVSR and Zero-AVSR using audio-only (A) and audio-visual (AV) inputs under different SNR noise levels.}
\label{fig:3}
\vspace{-0.5cm}
\end{figure}

\subsection{Noise-robustness Experiments}
By employing audio-visual speech inputs, we can achieve more robust noise performance in speech recognition compared to when we employ audio-only speech inputs. In this section, we analyze the performance of both Cascaded Zero-AVSR and Zero-AVSR by differing the acoustic noise levels from -5\,dB SNR to 15\,dB SNR. The acoustic noise is uniformly sampled among natural, babble, music, speech partitions from MUSAN~\cite{snyder2015musan}. The analysis results are shown in Fig.~\ref{fig:3}. It shows the average WER for all 9 languages. In both Cascaded Zero-AVSR and Zero-AVSR, the audio-only (A) models' performances are significantly degraded according to the noise become strong. However, we can confirm that the audio-visual (AV) models show robust performances over the different noise levels.

\begin{figure}[t]
\centering
\centerline{\includegraphics[width=8.5cm]{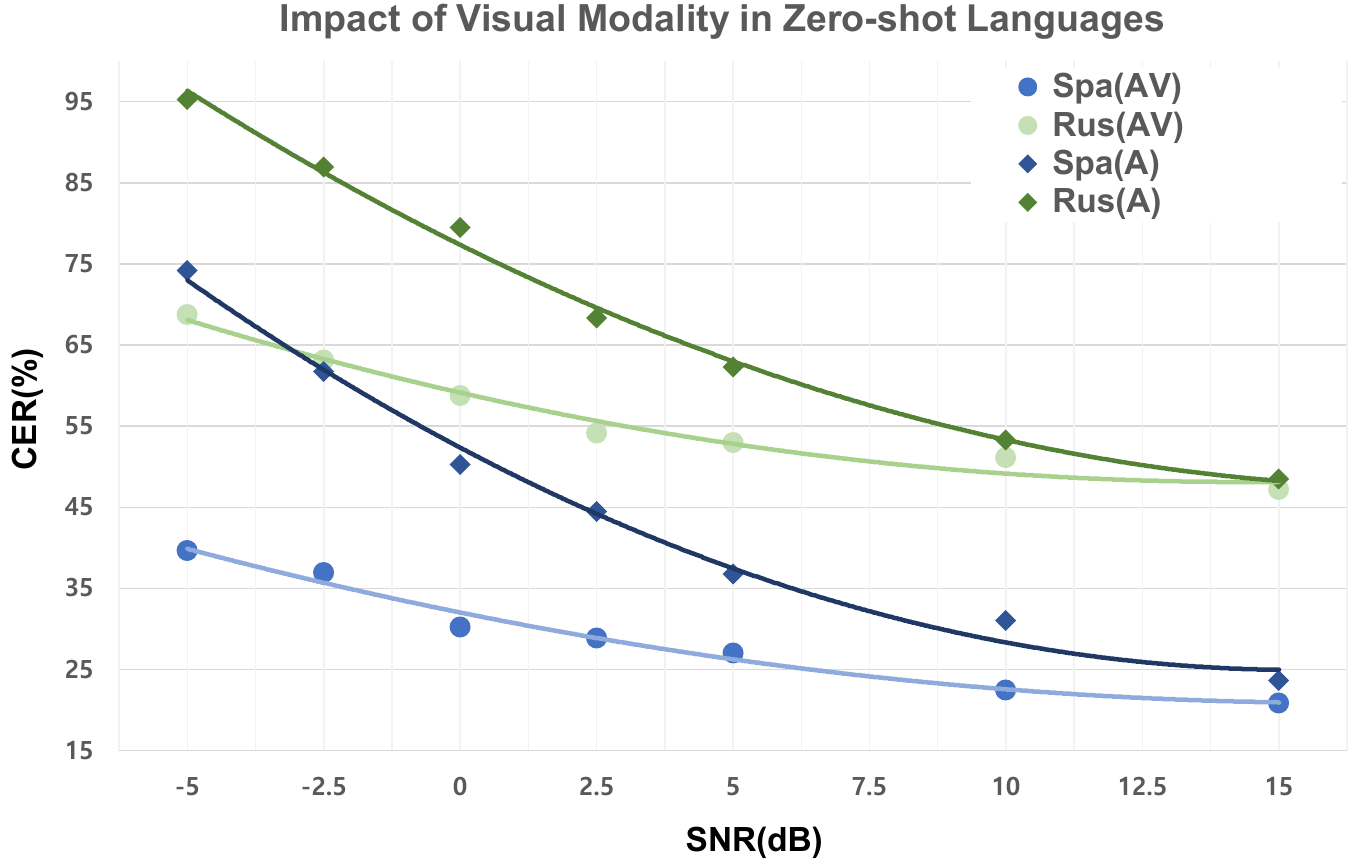}}
\vspace{-0.2cm}
\caption{Performance of Zero‑AVSR on zero‑shot languages, Spanish (Spa) and Russian (Rus), using audio‑only (A) and audio‑visual (AV) inputs across various SNR levels.}
\label{fig:5}
\vspace{-0.3cm}
\end{figure}
We have seen that by using audio-visual speech inputs, we can achieve more robust speech recognition performances. Here, we also analyze this in zero‑shot language settings. To this end, we measure CERs of Zero-AVSR model under various SNR levels on two unseen languages, Spanish and Russian. The results in Fig.~\ref{fig:5} show that, similar to the seen language scenarios, the audio-visual (AV) model outperforms the audio‑only (A) framework across all SNR levels. Especially, in the zero-shot language setting, we can observe that while both models exhibit comparable performance under clean environment (\ie, higher SNR), the performance gain by using audio-visual inputs over the audio‑only inputs increases as the SNR decreases.

\begin{figure}[t]
\centering
\centerline{\includegraphics[width=8.5cm]{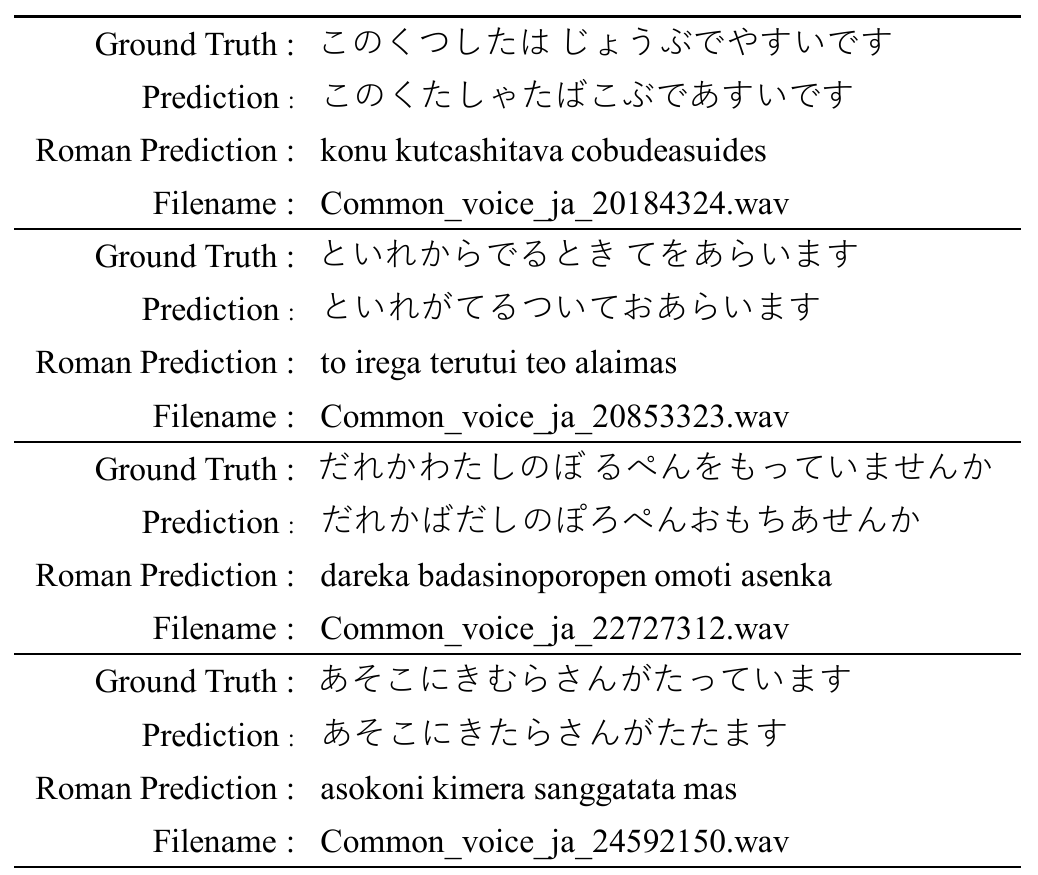}}
\vspace{-0.2cm}
\caption{Examples of prediction results from the Cascaded Zero-AVSR on an unseen language, Japanese, on out-of-domain data.}
\label{fig:4}
\vspace{-0.5cm}
\end{figure}
\subsection{Zero-Shot Performance on Out-of-Domain}
We evaluate the extent to which the proposed Zero-AVSR framework can perform on out-of-domain data. To this end, we evaluate the zero-shot speech recognition performance on Japanese, whose language family is not presented in the training set of MARC. We measure the Japanese performance on the test set of CommonVoice~\cite{ardila2019common} by using audio-only inputs. The Cascaded Zero-AVSR achieves 60.9\% CER and the Zero-AVSR achieves 64.9\% CER. These results demonstrate that the proposed Zero-AVSR framework can be employed for languages even when no data from the same language family is used, showing its scalability to more languages. The examples of prediction using the Cascaded Zero-AVSR are shown in Fig.~\ref{fig:4}. For example, as shown in the last row, the AV-Romanizer predicts the Roman text as `asokoni kimera sanggatata mas', and the LLM de-romanizes this text into Japanese as `\begin{CJK}{UTF8}{min}あそこにきたらさんがたたます\end{CJK}'. Notably, despite not being perfect, the prediction was made without Japanese data being employed during the training of the proposed AV-Romanizer. 

\subsection{Qualitative Error Analysis}
\begin{figure}[t]
\centering
\centerline{\includegraphics[width=8.5cm]{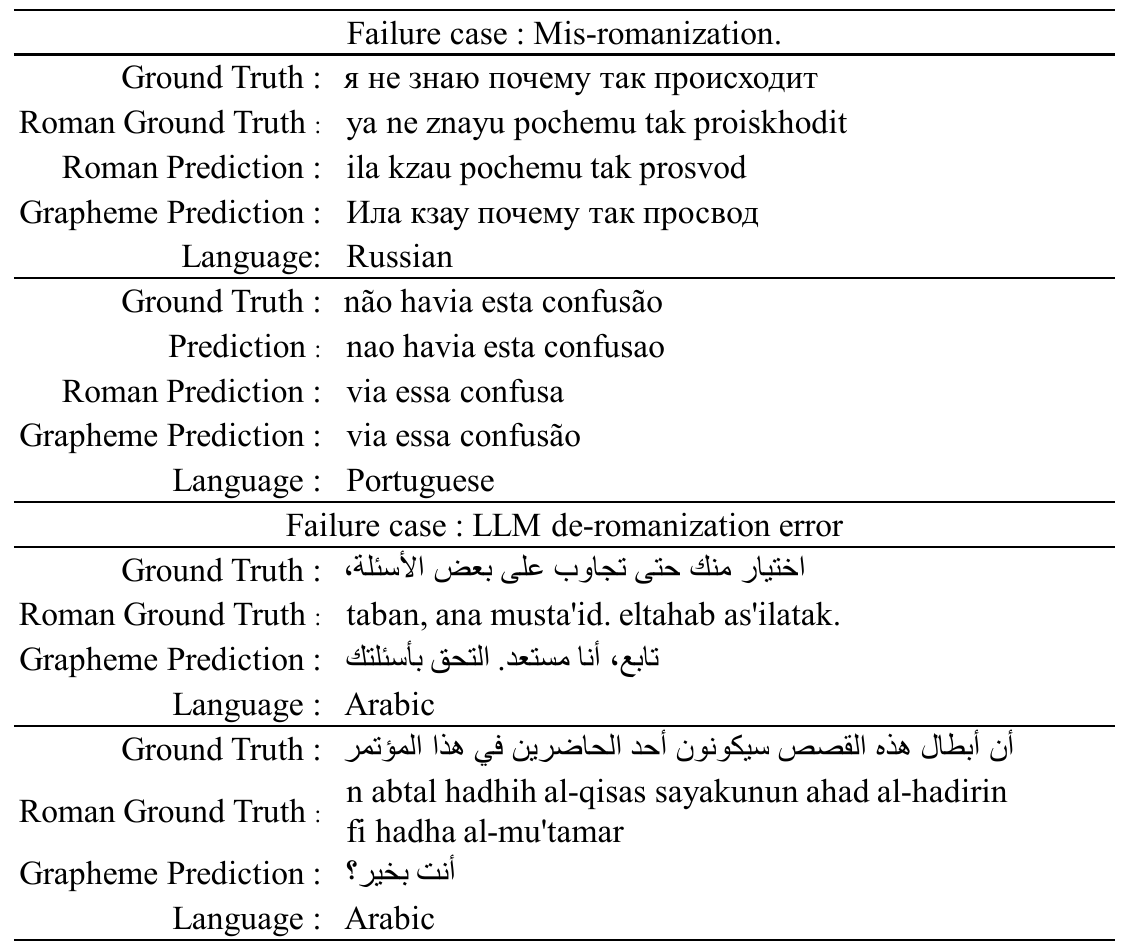}}
\vspace{-0.2cm}
\caption{Qualitative examples of mis‑romanization and LLM‑deromanization errors.}
\label{fig:6}
\vspace{-0.3cm}
\end{figure}

While our proposed Zero‑AVSR framework enables speech recognition in zero‑shot languages, its performance still lags behind that on seen languages. In this section, we conduct a qualitative error analysis to better understand the types and root causes of failures. Specifically, we investigate two error categories: mis‑romanization and LLM‑deromanization. Mis‑romanization errors occur when the AV‑romanizer’s output differs from the ground‑truth romanization. LLM‑deromanization errors arise when, after feeding the ground‑truth romanization into the LLM, the model mispredicts the original graphemes. Examples of both error types are illustrated in Fig.~\ref{fig:6}. In our analysis, we found that the vast majority of errors stem from mis‑romanization stage. This suggests that there is still room for improving the overall zero-shot language recognition performance by refining the romanization stage (\ie, the AV‑Romanizer).

\section{Detailed Information of MARC}
\label{sec:6}
The MARC dataset is driven by combining the existing audio-visual speech datasets. Specifically, the labeled audio-visual speech datasets, LRS3~\cite{afouras2018lrs3} and MuAViC~\cite{anwar2023muavic}, and the unlabeled audio-visual speech datasets, VoxCeleb2~\cite{chung2018voxceleb2} and AVSpeech~\cite{ephrat2018looking}, are combined. The information of each dataset is as follows:

{\bf Lip Reading Sentences 3 (LRS3)}~\cite{afouras2018lrs3} is a dataset designed for AVSR and is one of the most widely used resources. It contains 433 hours of audio-visual English data with human-annotated transcriptions, sourced from TED and TEDx talks.

{\bf Multilingual Audio-Visual Corpus (MuAViC)}~\cite{anwar2023muavic} is a dataset for multilingual audio-visual speech recognition and translation, collected from TED and TEDx talks across nine languages and comprising 1,200 hours of data with human-annotated transcriptions. Since its English portion overlaps with LRS3 and the preprocessing differs due to a different landmark detector, we exclusively use the English portion of LRS3 following the process in~\cite{ma2023auto}.

{\bf VoxCeleb2}~\cite{chung2018voxceleb2} is a dataset for speaker recognition containing 2,442 hours of multilingual audio-visual data. Although it includes speaker ID information, it lacks human-annotated text transcriptions and language labels.

{\bf AVSpeech}~\cite{ephrat2018looking} is a dataset aimed at isolating a target speaker's voice from mixed audio, sourced from YouTube videos and comprising 4,700 hours of multilingual audio-visual data. Like VoxCeleb2, it does not provide human-annotated text transcriptions or language labels.

The unlabeled audio-visual datasets, VoxCeleb2 and AVSpeech, are labeled using language identification and ASR. For language identification, we use the MMS-LID-1024 model\footnote{\url{https://huggingface.co/facebook/mms-lid-1024}} with a threshold of 0.95 for the confidence score. To generate language-specific graphemes, we utilize the pre-trained MMS-1B-ALL\footnote{\url{https://huggingface.co/facebook/mms-1b-all}} ASR model in conjunction with a language adapter, which is selected based on the identified language. Furthermore, during the decoding stage, we leverage language-specific language models\footnote{\url{https://huggingface.co/facebook/mms-cclms}}.
The resulting MARC dataset consists of 82 languages and approximately 2,916 hours of audio-visual data. The languages and their respective families~\cite{katzner2002languages} in the MARC dataset are listed in Tables~\ref{tab:MARC1} and \ref{tab:MARC2}.

\vspace{-0.1cm}
\section{Conclusion}
\vspace{-0.1cm}
We presented a zero-shot AVSR framework, Zero-AVSR, which extends language support beyond seen languages. The key idea behind this is 1) learning language-agnostic speech representations using Roman text, and 2) employing the LLMs to generate language-specific graphemes. To this end, we introduced the AV-Romanizer to convert audio-visual speech input into Roman text, which is then de-romanized into native scripts by LLMs. Depending on whether it involves fine-tuning an LLM, we explored two variants of frameworks, Cascaded Zero-AVSR and Zero-AVSR. Finally, we introduced the MARC dataset, comprising audio-visual speech data for 82 languages. Extensive experiments verified the effectiveness of Zero-AVSR.

\section*{Limitation}
Despite demonstrating strong zero‑shot performance, our framework exhibits two key limitations: 1) Language coverage depends on LLM support. Our AVSR pipeline relies on an LLM for each target language. When the LLM underperforms on a given language, particularly low‑resource ones, pipeline accuracy degrades accordingly. 2) Prosodic information is lost through romanization. We convert all inputs to unaccented Roman characters, which inherently discards tone, stress, and vowel‑length distinctions. This omission is especially critical in tonal languages, where pitch shifts can change word meaning entirely. To address these issues, future zero‑shot AVSR systems may consider 1) incorporating prosodic features, tone, stress, and rhythm into language‑agnostic encodings, and 2) leveraging next‑generation LLMs trained on low‑resource languages. These enhancements will broaden robust support across diverse linguistic contexts.

\section*{Acknowledgments}
This work was partly supported by two funds: the National Research Foundation of Korea (NRF) grant funded by the Korea government (MSIT) (No. NRF-2022R1A2C2005529), and Institute of Information \& communications Technology Planning \& Evaluation (IITP) grant funded by the Korea government(MSIT) (No.2022-0-00124, Development of Artificial Intelligence Technology for Self-Improving Competency-Aware Learning Capabilities).

{
    \small
    \bibliographystyle{unsrt}
    \bibliography{main}

\begin{thebibliography}{10}

\bibitem{afouras2018deep}
Triantafyllos Afouras, Joon~Son Chung, Andrew Senior, Oriol Vinyals, and Andrew Zisserman.
\newblock Deep audio-visual speech recognition.
\newblock {\em IEEE transactions on pattern analysis and machine intelligence}, 44(12):8717--8727, 2018.

\bibitem{petridis2018end}
Stavros Petridis, Themos Stafylakis, Pingehuan Ma, Feipeng Cai, Georgios Tzimiropoulos, and Maja Pantic.
\newblock End-to-end audiovisual speech recognition.
\newblock In {\em 2018 IEEE international conference on acoustics, speech and signal processing (ICASSP)}, pages 6548--6552. IEEE, 2018.

\bibitem{makino2019recurrent}
Takaki Makino, Hank Liao, Yannis Assael, Brendan Shillingford, Basilio Garcia, Otavio Braga, and Olivier Siohan.
\newblock Recurrent neural network transducer for audio-visual speech recognition.
\newblock In {\em 2019 IEEE automatic speech recognition and understanding workshop (ASRU)}, pages 905--912. IEEE, 2019.

\bibitem{ma2021end}
Pingchuan Ma, Stavros Petridis, and Maja Pantic.
\newblock End-to-end audio-visual speech recognition with conformers.
\newblock In {\em ICASSP 2021-2021 IEEE International Conference on Acoustics, Speech and Signal Processing (ICASSP)}, pages 7613--7617. IEEE, 2021.

\bibitem{ren2021learning}
Sucheng Ren, Yong Du, Jianming Lv, Guoqiang Han, and Shengfeng He.
\newblock Learning from the master: Distilling cross-modal advanced knowledge for lip reading.
\newblock In {\em Proceedings of the IEEE/CVF Conference on Computer Vision and Pattern Recognition}, pages 13325--13333, 2021.

\bibitem{shi2022learning}
Bowen Shi, Wei-Ning Hsu, Kushal Lakhotia, and Abdelrahman Mohamed.
\newblock Learning audio-visual speech representation by masked multimodal cluster prediction.
\newblock {\em arXiv preprint arXiv:2201.02184}, 2022.

\bibitem{hong2022visual}
Joanna Hong, Minsu Kim, Daehun Yoo, and Yong~Man Ro.
\newblock Visual context-driven audio feature enhancement for robust end-to-end audio-visual speech recognition.
\newblock In {\em Interspeech}, 2022.

\bibitem{serdyuk2022transformer}
Dmitriy Serdyuk, Otavio Braga, and Olivier Siohan.
\newblock Transformer-based video front-ends for audio-visual speech recognition for single and multi-person video.
\newblock {\em arXiv preprint arXiv:2201.10439}, 2022.

\bibitem{ma2023auto}
Pingchuan Ma, Alexandros Haliassos, Adriana Fernandez-Lopez, Honglie Chen, Stavros Petridis, and Maja Pantic.
\newblock Auto-avsr: Audio-visual speech recognition with automatic labels.
\newblock In {\em ICASSP 2023-2023 IEEE International Conference on Acoustics, Speech and Signal Processing (ICASSP)}, pages 1--5. IEEE, 2023.

\bibitem{hong2023watch}
Joanna Hong, Minsu Kim, Jeongsoo Choi, and Yong~Man Ro.
\newblock Watch or listen: Robust audio-visual speech recognition with visual corruption modeling and reliability scoring.
\newblock In {\em Proceedings of the IEEE/CVF Conference on Computer Vision and Pattern Recognition}, pages 18783--18794, 2023.

\bibitem{cappellazzo2024large}
Umberto Cappellazzo, Minsu Kim, Honglie Chen, Pingchuan Ma, Stavros Petridis, Daniele Falavigna, Alessio Brutti, and Maja Pantic.
\newblock Large language models are strong audio-visual speech recognition learners.
\newblock {\em arXiv preprint arXiv:2409.12319}, 2024.

\bibitem{rouditchenko2024whisper}
Andrew Rouditchenko, Yuan Gong, Samuel Thomas, Leonid Karlinsky, Hilde Kuehne, Rogerio Feris, and James Glass.
\newblock Whisper-flamingo: Integrating visual features into whisper for audio-visual speech recognition and translation.
\newblock {\em arXiv preprint arXiv:2406.10082}, 2024.

\bibitem{haliassos2024unified}
Alexandros Haliassos, Rodrigo Mira, Honglie Chen, Zoe Landgraf, Stavros Petridis, and Maja Pantic.
\newblock Unified speech recognition: A single model for auditory, visual, and audiovisual inputs.
\newblock In {\em The Thirty-eighth Annual Conference on Neural Information Processing Systems}, 2024.

\bibitem{amodei2016deep}
Dario Amodei, Sundaram Ananthanarayanan, Rishita Anubhai, Jingliang Bai, Eric Battenberg, Carl Case, Jared Casper, Bryan Catanzaro, Qiang Cheng, Guoliang Chen, et~al.
\newblock Deep speech 2: End-to-end speech recognition in english and mandarin.
\newblock In {\em International conference on machine learning}, pages 173--182. PMLR, 2016.

\bibitem{kim2017joint}
Suyoun Kim, Takaaki Hori, and Shinji Watanabe.
\newblock Joint ctc-attention based end-to-end speech recognition using multi-task learning.
\newblock In {\em 2017 IEEE international conference on acoustics, speech and signal processing (ICASSP)}, pages 4835--4839. IEEE, 2017.

\bibitem{prabhavalkar2023end}
Rohit Prabhavalkar, Takaaki Hori, Tara~N Sainath, Ralf Schl{\"u}ter, and Shinji Watanabe.
\newblock End-to-end speech recognition: A survey.
\newblock {\em IEEE/ACM Transactions on Audio, Speech, and Language Processing}, 32:325--351, 2023.

\bibitem{ma2022visual}
Pingchuan Ma, Stavros Petridis, and Maja Pantic.
\newblock Visual speech recognition for multiple languages in the wild.
\newblock {\em Nature Machine Intelligence}, 4(11):930--939, 2022.

\bibitem{haliassos2022jointly}
Alexandros Haliassos, Pingchuan Ma, Rodrigo Mira, Stavros Petridis, and Maja Pantic.
\newblock Jointly learning visual and auditory speech representations from raw data.
\newblock {\em arXiv preprint arXiv:2212.06246}, 2022.

\bibitem{kim2023lip}
Minsu Kim, Jeong~Hun Yeo, Jeongsoo Choi, and Yong~Man Ro.
\newblock Lip reading for low-resource languages by learning and combining general speech knowledge and language-specific knowledge.
\newblock In {\em Proceedings of the IEEE/CVF International Conference on Computer Vision}, pages 15359--15371, 2023.

\bibitem{kim2024efficient}
Minsu Kim, Jeonghun Yeo, Se~Jin Park, Hyeongseop Rha, and Yong~Man Ro.
\newblock Efficient training for multilingual visual speech recognition: Pre-training with discretized visual speech representation.
\newblock In {\em Proceedings of the 32nd ACM International Conference on Multimedia}, pages 1311--1320, 2024.

\bibitem{yeo2024visual}
Jeong~Hun Yeo, Seunghee Han, Minsu Kim, and Yong~Man Ro.
\newblock Where visual speech meets language: Vsp-llm framework for efficient and context-aware visual speech processing.
\newblock {\em arXiv preprint arXiv:2402.15151}, 2024.

\bibitem{yeo2024visual2}
Jeong~Hun Yeo, Minsu Kim, Shinji Watanabe, and Yong~Man Ro.
\newblock Visual speech recognition for languages with limited labeled data using automatic labels from whisper.
\newblock In {\em ICASSP 2024-2024 IEEE International Conference on Acoustics, Speech and Signal Processing (ICASSP)}, pages 10471--10475. IEEE, 2024.

\bibitem{haliassos2024braven}
Alexandros Haliassos, Andreas Zinonos, Rodrigo Mira, Stavros Petridis, and Maja Pantic.
\newblock Braven: Improving self-supervised pre-training for visual and auditory speech recognition.
\newblock In {\em ICASSP 2024-2024 IEEE International Conference on Acoustics, Speech and Signal Processing (ICASSP)}, pages 11431--11435. IEEE, 2024.

\bibitem{salesky2021multilingual}
Elizabeth Salesky, Matthew Wiesner, Jacob Bremerman, Roldano Cattoni, Matteo Negri, Marco Turchi, Douglas~W Oard, and Matt Post.
\newblock The multilingual tedx corpus for speech recognition and translation.
\newblock {\em arXiv preprint arXiv:2102.01757}, 2021.

\bibitem{anwar2023muavic}
Mohamed Anwar, Bowen Shi, Vedanuj Goswami, Wei-Ning Hsu, Juan Pino, and Changhan Wang.
\newblock Muavic: A multilingual audio-visual corpus for robust speech recognition and robust speech-to-text translation.
\newblock {\em arXiv preprint arXiv:2303.00628}, 2023.

\bibitem{zinonos2023learning}
Andreas Zinonos, Alexandros Haliassos, Pingchuan Ma, Stavros Petridis, and Maja Pantic.
\newblock Learning cross-lingual visual speech representations.
\newblock In {\em ICASSP 2023-2023 IEEE International Conference on Acoustics, Speech and Signal Processing (ICASSP)}, pages 1--5. IEEE, 2023.

\bibitem{han2024xlavs}
HyoJung Han, Mohamed Anwar, Juan Pino, Wei-Ning Hsu, Marine Carpuat, Bowen Shi, and Changhan Wang.
\newblock Xlavs-r: Cross-lingual audio-visual speech representation learning for noise-robust speech perception.
\newblock {\em arXiv preprint arXiv:2403.14402}, 2024.

\bibitem{burchi2024multilingual}
Maxime Burchi, Krishna~C Puvvada, Jagadeesh Balam, Boris Ginsburg, and Radu Timofte.
\newblock Multilingual audio-visual speech recognition with hybrid ctc/rnn-t fast conformer.
\newblock In {\em ICASSP 2024-2024 IEEE International Conference on Acoustics, Speech and Signal Processing (ICASSP)}, pages 10211--10215. IEEE, 2024.

\bibitem{touvron2023llama}
Hugo Touvron, Thibaut Lavril, Gautier Izacard, Xavier Martinet, Marie-Anne Lachaux, Timoth{\'e}e Lacroix, Baptiste Rozi{\`e}re, Naman Goyal, Eric Hambro, Faisal Azhar, et~al.
\newblock Llama: Open and efficient foundation language models.
\newblock {\em arXiv preprint arXiv:2302.13971}, 2023.

\bibitem{achiam2023gpt}
Josh Achiam, Steven Adler, Sandhini Agarwal, Lama Ahmad, Ilge Akkaya, Florencia~Leoni Aleman, Diogo Almeida, Janko Altenschmidt, Sam Altman, Shyamal Anadkat, et~al.
\newblock Gpt-4 technical report.
\newblock {\em arXiv preprint arXiv:2303.08774}, 2023.

\bibitem{jiang2023mistral}
Albert~Q Jiang, Alexandre Sablayrolles, Arthur Mensch, Chris Bamford, Devendra~Singh Chaplot, Diego de~las Casas, Florian Bressand, Gianna Lengyel, Guillaume Lample, Lucile Saulnier, et~al.
\newblock Mistral 7b.
\newblock {\em arXiv preprint arXiv:2310.06825}, 2023.

\bibitem{afouras2018lrs3}
Triantafyllos Afouras, Joon~Son Chung, and Andrew Zisserman.
\newblock Lrs3-ted: a large-scale dataset for visual speech recognition.
\newblock {\em arXiv preprint arXiv:1809.00496}, 2018.

\bibitem{chung2017lip}
Joon~Son Chung and Andrew Zisserman.
\newblock Lip reading in the wild.
\newblock In {\em Computer Vision--ACCV 2016: 13th Asian Conference on Computer Vision, Taipei, Taiwan, November 20-24, 2016, Revised Selected Papers, Part II 13}, pages 87--103. Springer, 2017.

\bibitem{vaswani2017attention}
A~Vaswani.
\newblock Attention is all you need.
\newblock {\em Advances in Neural Information Processing Systems}, 2017.

\bibitem{gulati2020conformer}
Anmol Gulati, James Qin, Chung-Cheng Chiu, Niki Parmar, Yu~Zhang, Jiahui Yu, Wei Han, Shibo Wang, Zhengdong Zhang, Yonghui Wu, et~al.
\newblock Conformer: Convolution-augmented transformer for speech recognition.
\newblock {\em arXiv preprint arXiv:2005.08100}, 2020.

\bibitem{hermjakob2018out}
Ulf Hermjakob, Jonathan May, and Kevin Knight.
\newblock Out-of-the-box universal romanization tool uroman.
\newblock In {\em Proceedings of ACL 2018, system demonstrations}, pages 13--18, 2018.

\bibitem{radford2023robust}
Alec Radford, Jong~Wook Kim, Tao Xu, Greg Brockman, Christine McLeavey, and Ilya Sutskever.
\newblock Robust speech recognition via large-scale weak supervision.
\newblock In {\em International conference on machine learning}, pages 28492--28518. PMLR, 2023.

\bibitem{pratap2024scaling}
Vineel Pratap, Andros Tjandra, Bowen Shi, Paden Tomasello, Arun Babu, Sayani Kundu, Ali Elkahky, Zhaoheng Ni, Apoorv Vyas, Maryam Fazel-Zarandi, et~al.
\newblock Scaling speech technology to 1,000+ languages.
\newblock {\em Journal of Machine Learning Research}, 25(97):1--52, 2024.

\bibitem{liu2018completely}
Da-Rong Liu, Kuan-Yu Chen, Hung-Yi Lee, and Lin-shan Lee.
\newblock Completely unsupervised phoneme recognition by adversarially learning mapping relationships from audio embeddings.
\newblock {\em arXiv preprint arXiv:1804.00316}, 2018.

\bibitem{chen2019completely}
Kuan-Yu Chen, Che-Ping Tsai, Da-Rong Liu, Hung-Yi Lee, and Lin-shan Lee.
\newblock Completely unsupervised speech recognition by a generative adversarial network harmonized with iteratively refined hidden markov models.
\newblock {\em arXiv preprint arXiv:1904.04100}, 2019.

\bibitem{baevski2021unsupervised}
Alexei Baevski, Wei-Ning Hsu, Alexis Conneau, and Michael Auli.
\newblock Unsupervised speech recognition.
\newblock {\em Advances in Neural Information Processing Systems}, 34, 2021.

\bibitem{li2022asr2k}
Xinjian Li, Florian Metze, David~R Mortensen, Alan~W Black, and Shinji Watanabe.
\newblock Asr2k: Speech recognition for around 2000 languages without audio.
\newblock {\em arXiv preprint arXiv:2209.02842}, 2022.

\bibitem{li2020universal}
Xinjian Li, Siddharth Dalmia, Juncheng Li, Matthew Lee, Patrick Littell, Jiali Yao, Antonios Anastasopoulos, David~R Mortensen, Graham Neubig, Alan~W Black, et~al.
\newblock Universal phone recognition with a multilingual allophone system.
\newblock In {\em ICASSP 2020-2020 IEEE International Conference on Acoustics, Speech and Signal Processing (ICASSP)}, pages 8249--8253. IEEE, 2020.

\bibitem{li2022zero}
Xinjian Li, Florian Metze, David~R Mortensen, Shinji Watanabe, and Alan~W Black.
\newblock Zero-shot learning for grapheme to phoneme conversion with language ensemble.
\newblock In {\em Findings of the Association for Computational Linguistics: ACL 2022}, pages 2106--2115, 2022.

\bibitem{zhao2024scaling}
Jinming Zhao, Vineel Pratap, and Michael Auli.
\newblock Scaling a simple approach to zero-shot speech recognition.
\newblock {\em arXiv preprint arXiv:2407.17852}, 2024.

\bibitem{radford2019language}
Alec Radford, Jeffrey Wu, Rewon Child, David Luan, Dario Amodei, Ilya Sutskever, et~al.
\newblock Language models are unsupervised multitask learners.
\newblock {\em OpenAI blog}, 1(8):9, 2019.

\bibitem{lakhotia2021generative}
Kushal Lakhotia, Eugene Kharitonov, Wei-Ning Hsu, Yossi Adi, Adam Polyak, Benjamin Bolte, Tu-Anh Nguyen, Jade Copet, Alexei Baevski, Abdelrahman Mohamed, et~al.
\newblock On generative spoken language modeling from raw audio.
\newblock {\em Transactions of the Association for Computational Linguistics}, 9:1336--1354, 2021.

\bibitem{latif2023sparks}
Siddique Latif, Moazzam Shoukat, Fahad Shamshad, Muhammad Usama, Yi~Ren, Heriberto Cuay{\'a}huitl, Wenwu Wang, Xulong Zhang, Roberto Togneri, Erik Cambria, et~al.
\newblock Sparks of large audio models: A survey and outlook.
\newblock {\em arXiv preprint arXiv:2308.12792}, 2023.

\bibitem{park2024lets}
Se~Jin Park, Chae~Won Kim, Hyeongseop Rha, Minsu Kim, Joanna Hong, Jeong~Hun Yeo, and Yong~Man Ro.
\newblock Let's go real talk: Spoken dialogue model for face-to-face conversation.
\newblock In {\em Proceedings of the 62nd Annual Meeting of the Association for Computational Linguistics (Volume 1: Long Papers)}, 2024.

\bibitem{hu2021lora}
Edward~J Hu, Yelong Shen, Phillip Wallis, Zeyuan Allen-Zhu, Yuanzhi Li, Shean Wang, Lu~Wang, and Weizhu Chen.
\newblock Lora: Low-rank adaptation of large language models.
\newblock {\em arXiv preprint arXiv:2106.09685}, 2021.

\bibitem{tang2023salmonn}
Changli Tang, Wenyi Yu, Guangzhi Sun, Xianzhao Chen, Tian Tan, Wei Li, Lu~Lu, Zejun Ma, and Chao Zhang.
\newblock Salmonn: Towards generic hearing abilities for large language models.
\newblock {\em arXiv preprint arXiv:2310.13289}, 2023.

\bibitem{chu2023qwen}
Yunfei Chu, Jin Xu, Xiaohuan Zhou, Qian Yang, Shiliang Zhang, Zhijie Yan, Chang Zhou, and Jingren Zhou.
\newblock Qwen-audio: Advancing universal audio understanding via unified large-scale audio-language models.
\newblock {\em arXiv preprint arXiv:2311.07919}, 2023.

\bibitem{dettmers2023qlora}
Tim Dettmers, Artidoro Pagnoni, Ari Holtzman, and Luke Zettlemoyer.
\newblock Qlora: Efficient finetuning of quantized llms.
\newblock {\em Advances in neural information processing systems}, 36:10088--10115, 2023.

\bibitem{schultz2001language}
Tanja Schultz and Alex Waibel.
\newblock Language-independent and language-adaptive acoustic modeling for speech recognition.
\newblock {\em Speech Communication}, 35(1-2):31--51, 2001.

\bibitem{vu2014multilingual}
Ngoc~Thang Vu, David Imseng, Daniel Povey, Petr Motlicek, Tanja Schultz, and Herv{\'e} Bourlard.
\newblock Multilingual deep neural network based acoustic modeling for rapid language adaptation.
\newblock In {\em 2014 IEEE international Conference on acoustics, speech and signal processing (ICASSP)}, pages 7639--7643. IEEE, 2014.

\bibitem{kim2024textless}
Minsu Kim, Jeongsoo Choi, Dahun Kim, and Yong~Man Ro.
\newblock Textless unit-to-unit training for many-to-many multilingual speech-to-speech translation.
\newblock {\em IEEE/ACM Transactions on Audio, Speech, and Language Processing}, 2024.

\bibitem{chung2018voxceleb2}
Joon~Son Chung, Arsha Nagrani, and Andrew Zisserman.
\newblock Voxceleb2: Deep speaker recognition.
\newblock {\em arXiv preprint arXiv:1806.05622}, 2018.

\bibitem{ephrat2018looking}
Ariel Ephrat, Inbar Mosseri, Oran Lang, Tali Dekel, Kevin Wilson, Avinatan Hassidim, William~T Freeman, and Michael Rubinstein.
\newblock Looking to listen at the cocktail party: A speaker-independent audio-visual model for speech separation.
\newblock {\em arXiv preprint arXiv:1804.03619}, 2018.

\bibitem{graves2006connectionist}
Alex Graves, Santiago Fern{\'a}ndez, Faustino Gomez, and J{\"u}rgen Schmidhuber.
\newblock Connectionist temporal classification: labelling unsegmented sequence data with recurrent neural networks.
\newblock In {\em Proceedings of the 23rd international conference on Machine learning}, pages 369--376, 2006.

\bibitem{chen2024llmasr}
CHEN CHEN, Ruizhe Li, Yuchen Hu, Sabato~Marco Siniscalchi, Pin-Yu Chen, EngSiong Chng, and Chao-Han~Huck Yang.
\newblock It's never too late: Fusing acoustic information into large language models for automatic speech recognition.
\newblock In {\em The Twelfth International Conference on Learning Representations}, 2024.

\bibitem{hu2024large}
Yuchen Hu, CHEN CHEN, Chao-Han~Huck Yang, Ruizhe Li, Chao Zhang, Pin-Yu Chen, and EngSiong Chng.
\newblock Large language models are efficient learners of noise-robust speech recognition.
\newblock In {\em The Twelfth International Conference on Learning Representations}, 2024.

\bibitem{hsu2022u}
Wei-Ning Hsu and Bowen Shi.
\newblock u-hubert: Unified mixed-modal speech pretraining and zero-shot transfer to unlabeled modality.
\newblock {\em Advances in Neural Information Processing Systems}, 35:21157--21170, 2022.

\bibitem{snyder2015musan}
David Snyder, Guoguo Chen, and Daniel Povey.
\newblock Musan: A music, speech, and noise corpus.
\newblock {\em arXiv preprint arXiv:1510.08484}, 2015.

\bibitem{katzner2002languages}
Kenneth Katzner and Kirk Miller.
\newblock {\em The languages of the world}.
\newblock Routledge, 2002.

\bibitem{ardila2019common}
Rosana Ardila, Megan Branson, Kelly Davis, Michael Henretty, Michael Kohler, Josh Meyer, Reuben Morais, Lindsay Saunders, Francis~M Tyers, and Gregor Weber.
\newblock Common voice: A massively-multilingual speech corpus.
\newblock {\em arXiv preprint arXiv:1912.06670}, 2019.

\end{thebibliography}
}

\begin{table*}[ht!]
  \renewcommand{\arraystretch}{1.2}
  \renewcommand{\tabcolsep}{2mm}
  \centering
  \resizebox{0.8\linewidth}{!}{
  \begin{tabular}{ccccccc}
    \toprule
    \textbf{Language Family}
    & \textbf{Subgroup}
    & \textbf{Branch}
    & \textbf{Number}
    & \textbf{Language Name}
    & \textbf{Language Code}
    & \textbf{Video Hours}
    \\
    \hline
    \multirow{50}{*}{\textbf{Indo-European}} 
     & \multirow{9}{*}{\textbf{Germanic}}  & \multirow{5}{*}{\textbf{Western}} & 1 & English & eng & 435.2  \\
     &  & & 2 & German & deu & 327.5  \\
     &  & & 3 & Dutch & nld & 72.5  \\
     &  & & 4 & Afrikaans & afr & 1.7  \\
     &  & & 5 & Luxembourgish & ltz & 1.9  \\
     \cmidrule(lr){3-7}
     &  & \multirow{4}{*}{\textbf{Northern}} & 6 & Swedish & swe & 19.3  \\
     &  & & 7 & Danish & dan & 18.1  \\
     &  & & 8 & Norwegian & nob & 2.8  \\
     &  & & 9 & Icelandic & isl & 0.4  \\
     \cmidrule(lr){2-7}
     & \multirow{9}{*}{\textbf{Romance}} & \multirow{9}{*}{\textbf{-}} & 10 & Italian & ita & 146.8  \\
     &  & & 11 & French & fra & 291.7  \\
     &  & & 12 & Spanish & spa & 216.2  \\
     &  & & 13 & Portuguese & por & 408.0  \\
     &  & & 14 & Romanian & ron & 16.0  \\
     &  & & 15 & Catalan & cat & 7.0  \\
     &  & & 16 & Galician & glg & 8.7  \\
     &  & & 17 & Asturian & ast & 0.1  \\
     &  & & 18 & Occitan & oci & 1.5  \\
     \cmidrule(lr){2-7}
     & \multirow{2}{*}{\textbf{Celtic}} & \textbf{Brythonic} & 19 & Welsh & cym & 97.1  \\
     \cmidrule(lr){3-7}
     &  & \textbf{Goidelic} & 20 & Irish & gle & 0.1  \\
     \cmidrule(lr){2-7}
     & \textbf{Hellenic} & - & 21 & Greek & ell & 21.5  \\
     \cmidrule(lr){2-7}
     & \multirow{12}{*}{\textbf{Slavic}} & \multirow{3}{*}{\textbf{Eastern}} & 22 & Russian & rus & 123.8  \\
     &  & & 23 & Ukrainian & ukr & 3.8  \\
     &  & & 24 & Belarusian & bel & 5.9  \\
     \cmidrule(lr){3-7}
\textbf{}     &  & \multirow{3}{*}{\textbf{Western}} & 25 & Polish & pol & 50.5  \\
     &  & & 26 & Czech & ces & 20.6  \\
     &  & & 27 & Slovak & slk & 4.9  \\
     \cmidrule(lr){3-7}
     &  & \multirow{6}{*}{\textbf{Southern}} & 28 & Bulgarian & bul & 3.9  \\
     &  & & 29 & Slovene & slv & 4.6  \\
     &  & & 30 & Macedonian & mkd & 0.3  \\
     &  & & 31 & Bosnian & bos & 0.8  \\
     &  & & 32 & Croatian & hrv & 2.6  \\
     &  & & 33 & Serbian & srp & 0.9  \\
     \cmidrule(lr){2-7}
     & \multirow{14}{*}{\textbf{Indo-Iranian}} & \multirow{4}{*}{\textbf{Iranian}} & 34 & Persian & fas & 7.6  \\
     &  & & 35 & Kurdish & ckb & 0.05  \\
     &  & & 36 & Tajik & tgk & 0.1  \\
     &  & & 37 & Pushto  & pus & 0.3  \\
     \cmidrule(lr){3-7}
     &  & \multirow{10}{*}{\textbf{Indic}} & 38 & Hindi & hin & 99.0  \\
     &  & & 39 & Urdu & urd & 8.6  \\
     &  & & 40 & Bengali & ben & 8.8  \\
     &  & & 41 & Punjabi & pan & 3.0  \\
     &  & & 42 & Marathi & mar & 8.2  \\
     &  & & 43 & Gujarati & guj & 1.6  \\
     &  & & 44 & Assamese & asm & 0.2  \\
     &  & & 45 & Nepali & npi & 4.5  \\
     &  & & 46 & Sindhi & snd & 0.5  \\
     &  & & 47 & Odia & ory & 0.1  \\
    \bottomrule
  \end{tabular}}
  \caption{The Data Statistics of the MARC dataset 1.}
  \label{tab:MARC1}
\end{table*}

\begin{table*}[ht!]
  \renewcommand{\arraystretch}{1.2}
  \renewcommand{\tabcolsep}{2mm}
  \centering
  \resizebox{0.8\linewidth}{!}{
  \begin{tabular}{ccccccc}
    \toprule
    \textbf{Language Family}
    & \textbf{Subgroup}
    & \textbf{Branch}
    & \textbf{Number}
    & \textbf{Language Name}
    & \textbf{Language Code}
    & \textbf{Video Hours}
    \\
    \hline
    
    \multirow{3}{*}{\textbf{Indo-European}} & \multirow{2}{*}{\textbf{Baltic}} & - & 48 & Lithuanian & lit & 2.9  \\
     &  & - & 49 & Latvian & lav & 1.3  \\
     \cmidrule(lr){2-7}
     & - & -& 50 & Armenian & hye & 0.7  \\
     \cmidrule(lr){1-7}
    \multirow{3}{*}{\textbf{Uralic}} & \multirow{3}{*}{\textbf{Finno-Ugric}} & \multirow{2}{*}{\textbf{Finnic}} & 51 & Finnish & fin & 9.7  \\
     &  & & 52 & Estonian & est & 2.1  \\
     \cmidrule(lr){3-7}
     &  & \textbf{Ugric} & 53 & Hungarian & hun & 11.0  \\
     \cmidrule(lr){1-7}
    \multirow{6}{*}{\textbf{Altaic}} & \multirow{5}{*}{\textbf{Turkic}} & \multirow{2}{*}{\textbf{Southwestern}} & 54 & Turkish & tur & 50.6  \\
     &  & & 55 & Azerbaijani & aze & 1.9  \\
     \cmidrule(lr){3-7}
     &  & \multirow{2}{*}{\textbf{Northwestern}} & 56 & Kazakh & kaz & 3.8  \\
     &  & & 57 & Kyrgyz & kir & 0.1  \\
     \cmidrule(lr){3-7}
     &  & \textbf{Southeastern} & 58 & Uzbek & uzb & 0.3  \\
     \cmidrule(lr){2-7}
     & \textbf{Mongolian} & - & 59 & Mongolian & mon & 1.1  \\
    \cmidrule(lr){1-7}
    \textbf{Caucasian} & \textbf{Southern} & - & 60 & Georgian & kat & 1.5  \\
    \cmidrule(lr){1-7}
    \multirow{4}{*}{\textbf{Dravidian}}  & - & - & 61 & Telugu & tel & 18.8  \\
         & -     & - & 62 & Tamil   & tam & 18.3 \\
         & -     & - & 63 & Kannada & kan & 3.6 \\
         & -     & - & 64 & Malayalam   & mal & 15.5 \\
    \cmidrule(lr){1-7}
    \textbf{Independent}   & -     & - & 65 & Korean  & kor & 128.6 \\
    \cmidrule(lr){1-7}
    \textbf{Mon-Khmer}     & -     & - & 66 & Vietnamese  & vie & 32.9 \\
    \cmidrule(lr){1-7}
    \multirow{4}{*}{\textbf{Austronesian}} & \multirow{3}{*}{\textbf{Western}}     & - & 67 & Indonesian  & ind & 15.0 \\
      &      & - & 68 & Javanese    & jav & 0.1 \\
      &      & - & 69 & Tagalog     & tgl & 6.1 \\
      \cmidrule(lr){2-7}
      & \textbf{Polynesian}  & - & 70 & Maori  & mri & 4.9 \\
    \cmidrule(lr){1-7}
    \multirow{5}{*}{\textbf{Niger-Congo}}   & \textbf{Atlantic}    & - & 71 & Wolof  & wol & 1.1 \\
    \cmidrule(lr){2-7}
      & \multirow{4}{*}{\textbf{Benue-Congo}}  & - & 72 & Swahili  & swh & 1.2 \\
      &  & - & 73 & Lingala  & lin & 0.9 \\
      &  & - & 74 & Ganda    & lug & 0.04 \\
      &  & - & 75 & Shona    & sna & 3.8 \\
    \cmidrule(lr){1-7}
    \multirow{7}{*}{\textbf{Afro-Asiatic}}  & \multirow{4}{*}{\textbf{Semitic}}  & \multirow{2}{*}{\textbf{North Arabic}} & 76 & Arabic  & ara & 96.7 \\
      &   &  & 77 & Maltese & mlt & 1.5 \\
      \cmidrule(lr){3-7}
      &   & \textbf{Canaanitic}   & 78 & Hebrew  & heb & 14.3 \\
      \cmidrule(lr){3-7}
      &   & \textbf{Ethiopic}     & 79 & Amharic & amh & 1.2  \\
      \cmidrule(lr){2-7}
      & \multirow{2}{*}{\textbf{Cushitic}}  & - & 80 & Somali   & som & 4.6  \\
      &  & - & 81 & Oromo    & orm & 0.1  \\
      \cmidrule(lr){2-7}
      & \textbf{Chadic}   & - & 82 & Hausa    & hau & 0.7  \\
    \bottomrule
  \end{tabular}}
  \caption{The Data Statistics of the MARC dataset 2.}
  \label{tab:MARC2}
\end{table*}


\end{document}